\documentclass[times, review, 10pt]{elsarticle}

\usepackage{amssymb}
\usepackage{amsmath}
\usepackage{color}
\usepackage{tabularx}
\usepackage{multirow}

\usepackage{makecell}
\usepackage{caption}
\journal{PATTERN RECOGNITION}

\usepackage[switch]{lineno} 

\begin{document}

\begin{frontmatter}

\title{IFShip: Interpretable Fine-grained Ship Classification with Domain Knowledge-Enhanced Vision-Language Models}

\author[a]{Mingning Guo}

\author[a]{Mengwei Wu}

\author[a]{Yuxiang Shen}

\author[a]{Haifeng Li}

\author[a]{Chao Tao \corref{cor1}}
\ead{kingtaochao@csu.edu.cn}

\cortext[cor1]{Corresponding author}

\affiliation[a]{organization={School of Geosciences and Info-Physics, Central South University},
            city={Changsha},
            postcode={410083}, 
            country={China}}

\begin{abstract}
End-to-end interpretation currently dominates the remote sensing fine-grained ship classification (RS-FGSC) task. However, the inference process remains uninterpretable, leading to criticisms of these models as “black box” systems. To address this issue, we propose a domain knowledge-enhanced Chain-of-Thought (CoT) prompt generation mechanism, which is used to semi-automatically construct a task-specific instruction-following dataset, TITANIC-FGS. By training on TITANIC-FGS, we adapt general-domain vision-language models (VLMs) to the FGSC task, resulting in a model named IFShip. Building upon IFShip, we develop an FGSC visual chatbot that redefines the FGSC problem as a step-by-step reasoning task and conveys the reasoning process in natural language. Experimental results show that IFShip outperforms state-of-the-art FGSC algorithms in both interpretability and classification accuracy. Furthermore, compared to VLMs such as LLaVA and MiniGPT-4, IFShip demonstrates superior performance on the FGSC task. It provides an accurate chain of reasoning when fine-grained ship types are recognizable to the human eye and offers interpretable explanations when they are not. Our dataset is publicly available at: https://github.com/lostwolves/IFShip.
\end{abstract}

\begin{highlights}
\item We design a domain knowledge-enhanced Chain-of-Thought prompt generation mechanism.
\item We build an instruction-following dataset for fine-grained ship classification tasks.
\item We train a vision-language model for fine-grained ship classification named IFShip.
\end{highlights}


\end{frontmatter}

\section{Introduction}
\label{sec1}
Classifying fine-grained categories of ships, such as cargo ships, aircraft carriers, and destroyers, is crucial for analyzing their operational intentions \cite{cite4,cite1}. However, the complexity of this task arises from the subtle differences among ship categories, making it more challenging than distinguishing ships from non-ship entities  \cite{cite2}.

\begin{figure*}[ht]
  \centering
   \includegraphics[width=1\linewidth]{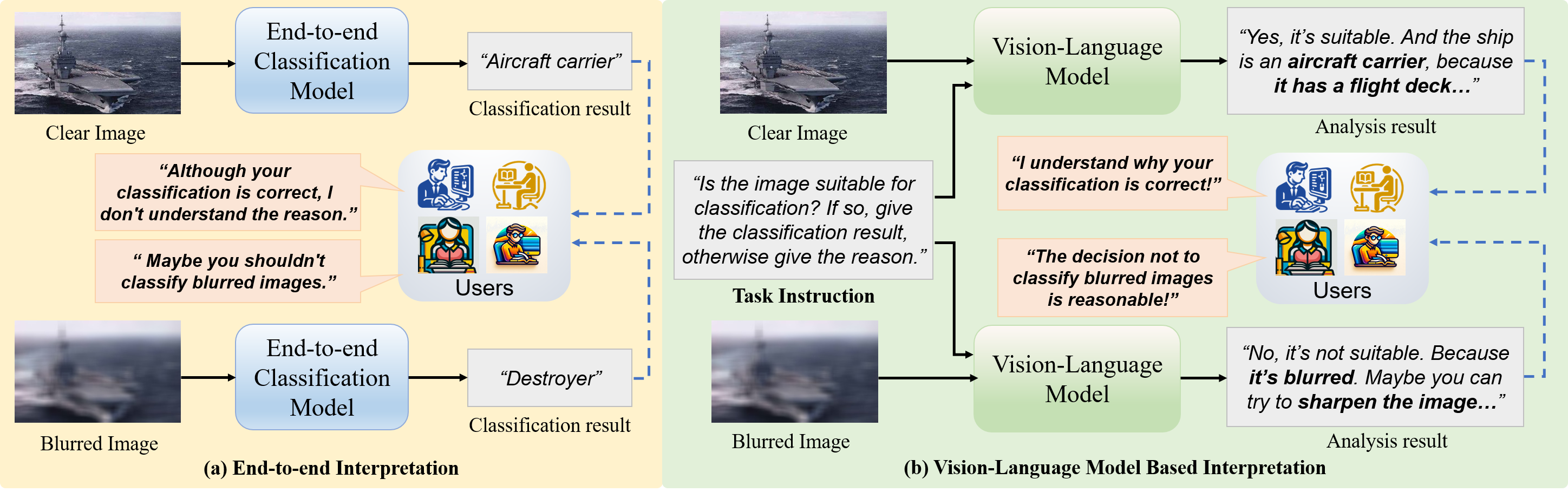}
   \caption{Comparison between traditional end-to-end interpretation and large vision-language model based interpretation.}
   \label{fig:com}
\end{figure*}

Currently, end-to-end interpretation \cite{cite6} is the main paradigm for remote sensing fine-grained ship classification (RS-FGSC) task. It uses a data-driven approach with attention mechanisms \cite{cite7,cite12} to automatically learn discriminative fine-grained ship features. This policy assigns varying weights to the learned features, allowing the model to prioritize important ones. For example, Yang et al. \cite{cite9} presents an adaptive mid-level feature attention learning approach for FGSC in optical remote sensing images. By introducing mid-level channel attention, the method identifies discriminative regions and subtle visual features, improving classification accuracy without additional annotations. However, end-to-end interpretation can only provide classification results without explaining the rationale behind its decision-making, as shown in \textcolor{blue}{Fig. \ref{fig:com} (a)}. While visualization tools such as CAM and Grad-CAM are available, they only highlight the image regions the model focuses on and do not provide direct insights into the model's decisions \cite{cite33}, resulting in limited interpretability. Moreover, the end-to-end interpretation methods forcibly provide classification results for all input images, even when the images are so blurred that their fine-grained categories are indistinguishable to the human eye. We argue that a truly intelligent interpretation model should provide an accurate chain of reasoning when fine-grained ship types are recognizable to the human eye and offer interpretable reasons when they are not, as shown in \textcolor{blue}{Fig. \ref{fig:com} (b)}.

In recent years, the advent of Vision-Language Models (VLMs) \cite{cite13} has made it possible to describe the image content in human language. For example, VLMs like LLaVA \cite{cite18} have demonstrated strong performance in providing explanations for open-world visual understanding tasks, such as classification and detection. This paradigm further extends from the general domain to biomedical scenarios, LLaVA-Med \cite{cite19} exhibit robust multimodal dialogue capabilities, offering detailed explanations for biomedical image-related queries. Similarly, RS domain-specific VLMs such as RSGPT \cite{cite20} and EarthGPT \cite{cite21} demonstrate interpretability in remote sensing tasks such as image description and visual grounding. The success of VLMs is attributed to the use of instruction-following datasets, which enables the model to align with human intent to effectively complete various real-world tasks. However, existing general remote sensing task instruction-following datasets are insufficient to support the FGSC task for the following reasons: 1) The FGSC task requires the model to focus on the fine-grained features of a ship, such as bow shape and ship island position. However, such details are usually not emphasized in instruction-following datasets designed for general remote sensing tasks. 2) Fine-grained ship features are highly sensitive to imaging conditions, requiring the model to adapt accordingly. However, existing instruction-following datasets in remote sensing lack domain-specific knowledge, hindering the model's ability to flexibly perform FGSC tasks.

To address these issues, we design a domain knowledge-enhanced Chain-of-Thought (CoT) \cite{cite50} prompt generation mechanism. By constructing the domain knowledge base and designing CoT instruction generation principles, this mechanism ensures that the generated multi-round dialogue instructions follow human-like logic for task reasoning. Using this mechanism, we construct the FGSC task-specific instruction-following dataset named TITANIC-FGS (Tactical Instruction Tuning for Accurate, Nimble, and Interpretable Classification of Fine-Grained Ships). The TITANIC-FGS dataset can guide VLMs to learn how to distinguish fine-grained ship types under various imaging conditions. Through training on TITANIC-FGS, we develop IFShip, a specialized VLM for FGSC built on the LLaVA framework, which can convey the reasoning process of its FGSC results in human language. Experimental results show that the FGSC visual chatbot based on IFShip outperforms state-of-the-art (SOTA) FGSC algorithms in both classification interpretability and accuracy. Furthermore, IFShip outperforms VLMs such as LLaVA and MiniGPT-4 in the FGSC problem reasoning task. It excels in providing a clear chain of reasoning when fine-grained ship types are recognizable to the human eye, and offers interpretable explanations when they are not. The main contributions of this paper are summarized as follows:

\begin{itemize}

\item We introduce a domain knowledge-enhanced CoT prompt generation mechanism. By incorporating domain knowledge and CoT generation principles into the instruction generation process, the instructions follow human-like logic for task reasoning. This mechanism holds the potential for general applicability across various tasks.
\item We present the first domain knowledge-enhanced instruction-following dataset for the FGSC task, TITANIC-FGS. This dataset is carefully designed to mimic human-like decision-making processes, enabling VLMs trained on it to perform FGSC tasks with reasoning aligned more closely with human logic.
\item We develop IFShip, a task-specific VLM for FGSC built on the LLaVA framework. Based on IFShip, we develop an FGSC visual chatbot that overcomes the interpretability limitations of traditional end-to-end classification methods. Compared to general VLMs such as LLaVA and MiniGPT-4, IFShip shows improved performance by avoiding incorrect answers and hallucinations.
\end{itemize}

\section{Related work}
\label{sec2}
\subsection{End-to-end Interpretation of FGSC}
\label{subsec2-1}
End-to-end interpretation refers to obtain classification results directly from the image inputs, without intermediate steps or manual intervention \cite{cite29}. In applying this paradigm to the FGSC task, the core challenge is how to obtain discriminative fine-grained ship features  \cite{cite8}. Early ship classification algorithms were mainly based on hand-crafted features \cite{cite63} like the length of the hull. However, these methods perform poorly on the FGSC task, which requires high-level details such as bow shape and ship island position. In contrast, convolutional neural networks (CNNs) \cite{cite15} learn image features in a data-driven manner, capturing both high-level and fine-grained details with convolutional kernels of varying sizes. For example, Huang et al. \cite{cite31} combined multiple low-level local CNN features with high-level global CNN features, achieving a breakthrough in precision over traditional methods in the FGSC task. Huang et al. \cite{cite66} developed a parallel network that integrates CNN and transformer architectures to extract ship features at various scales. Compared with the method based on hand-crafted features, these models provide more discriminative fine-grained ship features.

However, the features learned by these models are often unstable, and frequently focusing on background elements that are unrelated to the ship targets. Consequently, subsequent research has attempted to incorporate attention mechanisms into the model to help stabilize the focus on the detailed features. For example, Zhang et al. \cite{cite5} reweighted regional features through a recursive neural network attention mechanism, enabling the network to focus more on areas critical to FGSC. Gao et al. \cite{cite2} proposed a lightweight adaptive task attention mechanism that reduces interference from complex backgrounds by generating three-dimensional weights. Xiong et al. \cite{cite33} employed a multi-head attention mechanism integrated with a structural causal model to generate causal attention maps. These methods help the network focus on features that have a decisive impact on the final FGSC results through multiple weighting.

Nevertheless, these works have not addressed the lack of interpretability in the end-to-end interpretation paradigm. The “black box" nature of deep networks makes the decision-making process of classification models hard to interpret. Therefore, developing an interpretable FGSC method continues to be a significant challenge.

\subsection{Large Vision-Language Models}
\label{subsec2-2}
By integrating Large Language Models (LLMs) with visual encoders, the VLMs can leverage language to enhance the model's interpretability. This integration facilitates the understanding and explanation of visual data through linguistic descriptions, making complex visual content more accessible to humans \cite{cite3}.

In the field of computer vision, classic VLMs like VisualGPT \cite{cite39} have shown significant potential to address multimodal challenges, such as visual question answering (VQA). VLMs excel in these tasks due to training on large-scale instruction-following datasets. However, these datasets are generally derived from diverse internet sources, which makes it difficult for VLMs to adapt to domain-specific knowledge and task requirements. This underscores the necessity for domain-specific training datasets. In the biomedical field, Li et al. \cite{cite19} developed a biomedical image instruction-following dataset based on medical images and corresponding descriptions. Building on this, they trained a specialized model named LLaVA-Med, which outperforms general-domain VLMs in various biomedical visual question-answering tasks. In the field of autonomous driving, Xu et al. \cite{cite68} developed a visual instruction tuning dataset for explainable autonomous driving. On this dataset, they trained a model called DriveGPT4 for explainable end-to-end autonomous driving, enabling it to respond to human inquiries about the vehicle.

In the RS field, researchers have also begun to develop VLMs specifically for RS tasks. For example, Hu et al. \cite{cite20} created the RSICap dataset by annotating RS images with detailed descriptions of remote sensing scenes and objects, then trained the RSGPT model. Experiments on various RS tasks show that RSGPT outperforms other VLMs in extracting RS image information and enhancing VQA capabilities. Zhang et al. \cite{cite21} proposed the EarthGPT model, trained on the MMRS-1M dataset, which includes over 1 million image-text pairs from diverse RS sources. The training enables EarthGPT to excel in understanding optical, infrared, and SAR images, demonstrating superior multimodal comprehension compared to other VLMs. Kuckreja et al. \cite{cite69} expanded image-text pairs from existing diverse remote sensing datasets to generate a RS multimodal instruction-following dataset. The trained GeoChat model can not only respond to image-level queries but also accept dialogue opinions for specific regions.

While these efforts have advanced VLM applications in RS, these models perform poorly on the FGSC task. Because the FGSC task requires a focus on fine-grained ship features like parts and size, which are often overlooked in multi-task hybrid instruction-following datasets. Therefore, constructing a domain knowledge-enhanced dataset for FGSC is essential for training a VLM that excels in this task.

\section{Methodology}
\label{sec3}

\subsection{Why we need domain knowledge-enhanced instruction-following dataset}
\label{sec3-1}
Prompts with rich domain knowledge play an important role in constructing task-specific instruction-following datasets  \cite{cite10, cite11}. However, existing instruction-following datasets are generally constructed using a method based on prompt-template filling, where the prompts only offer basic task descriptions and lack detailed guidance on task-specific information. Additionally, the instruction generation rules in the prompts only provide the basic format of the instructions and do not design task-specific instruction formats. Consequently, the generated instructions do not effectively help VLMs capture the key information for specific tasks. Therefore, VLMs trained on these datasets often exhibit lower performance compared to traditional task-specific expert models.

\begin{figure*}[ht]
  \centering
   \includegraphics[width=1\linewidth]{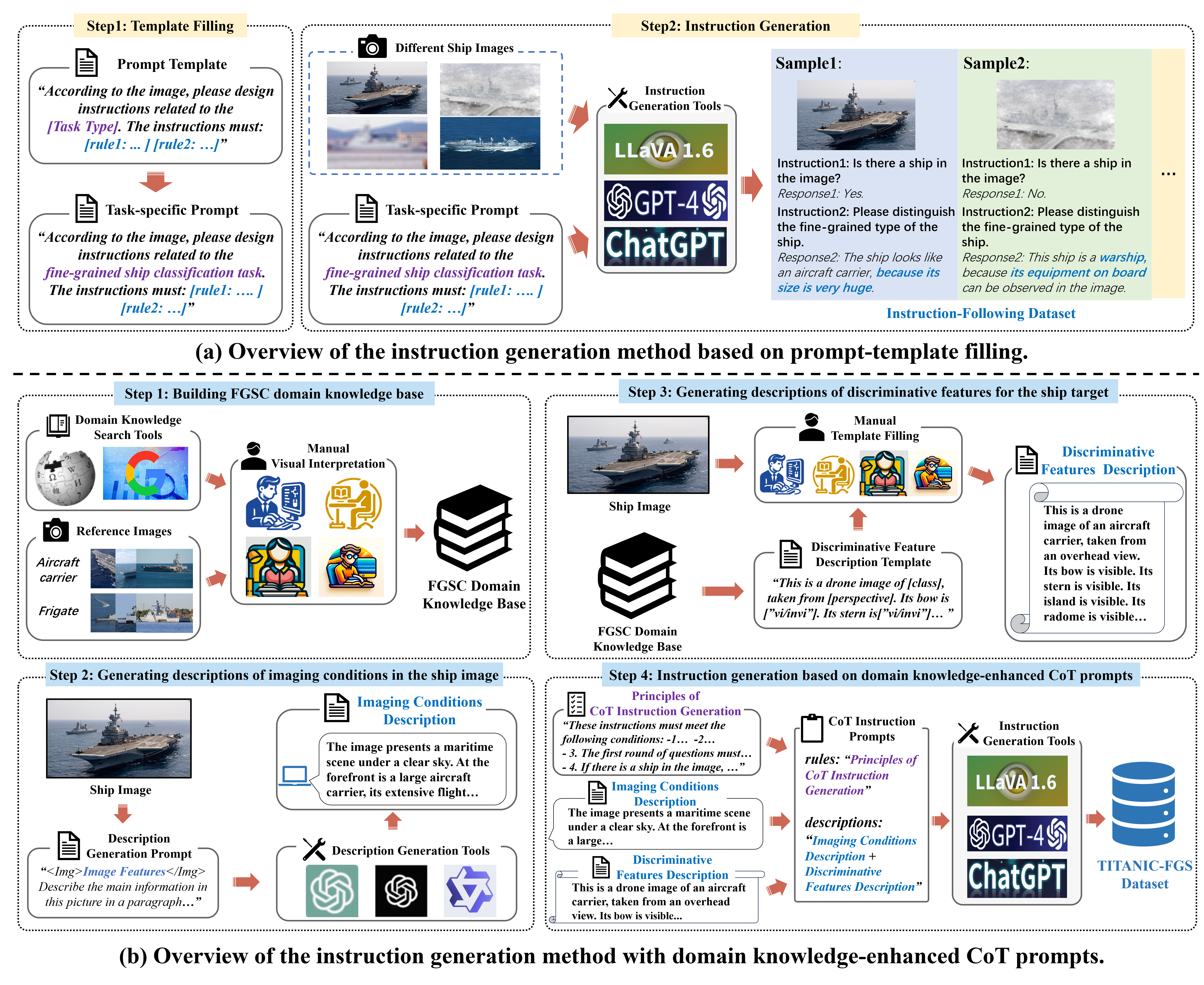}
   \caption{Comparison of Instruction Generation Methods: Prompt-Template Filling vs. Domain Knowledge-Enhanced CoT Prompts.}
   \label{fig:method compared}
\end{figure*}

Taking the FGSC task as an example, as illustrated in \textcolor{blue}{Fig. \ref{fig:method compared} (a)}, traditional prompt-template filling based method firstly designs a prompt template that contains only basic generation rules (\textit{“According to the image, please design instructions related to the [Task Type]. The instructions must: [rule1: …] [rule2: …]”}), and then fills in specific task types (\textit{“fine-grained ship classification task”}), creating the task-specific prompt. Then, the task-specific prompt and different task images are fed into the instruction generation tools (such as GPT-4) to construct the task-specific instruction-following dataset. The generated instruction-following dataset has two main issues: Firstly, the answers provided in the instruction dialogues fail to focus on specific details, such as key components, which makes it difficult to guarantee the accuracy of the dialogue. For example, the classification rationale in Sample 1, such as “its size is very huge” fails to support the “aircraft carrier” classification. Secondly, the multi-turn dialogues lack logical consistency and fail to guide VLMs in learning the correct chain of reasoning for the FGSC task. For example, the first dialogue round in Sample 2 confirms no ship targets, yet the second round asks about ship types.

Thus, to develop a domain knowledge-enhanced instruction-following dataset, we believe that the instruction generation prompts should have the following characteristics: Firstly, the prompts should include domain-specific information pertinent to the FGSC task. This allows the generated instructions to help VLMs in understanding the relationship between imaging conditions and discriminative ship features, thereby ensuring accurate classification results. Secondly, the prompts should include guidance that accounts for the correct logical relationships across multiple dialogue rounds. This allows VLMs to learn the specific reasoning chain required for the FGSC task, thereby enhancing the interpretability of the results.

\subsection{The construction of domain knowledge-enhanced instruction-following dataset: TITANIC-FGS}
\label{sec3-2}
\begin{figure*}[ht]
  \centering
   \includegraphics[width=1\linewidth]{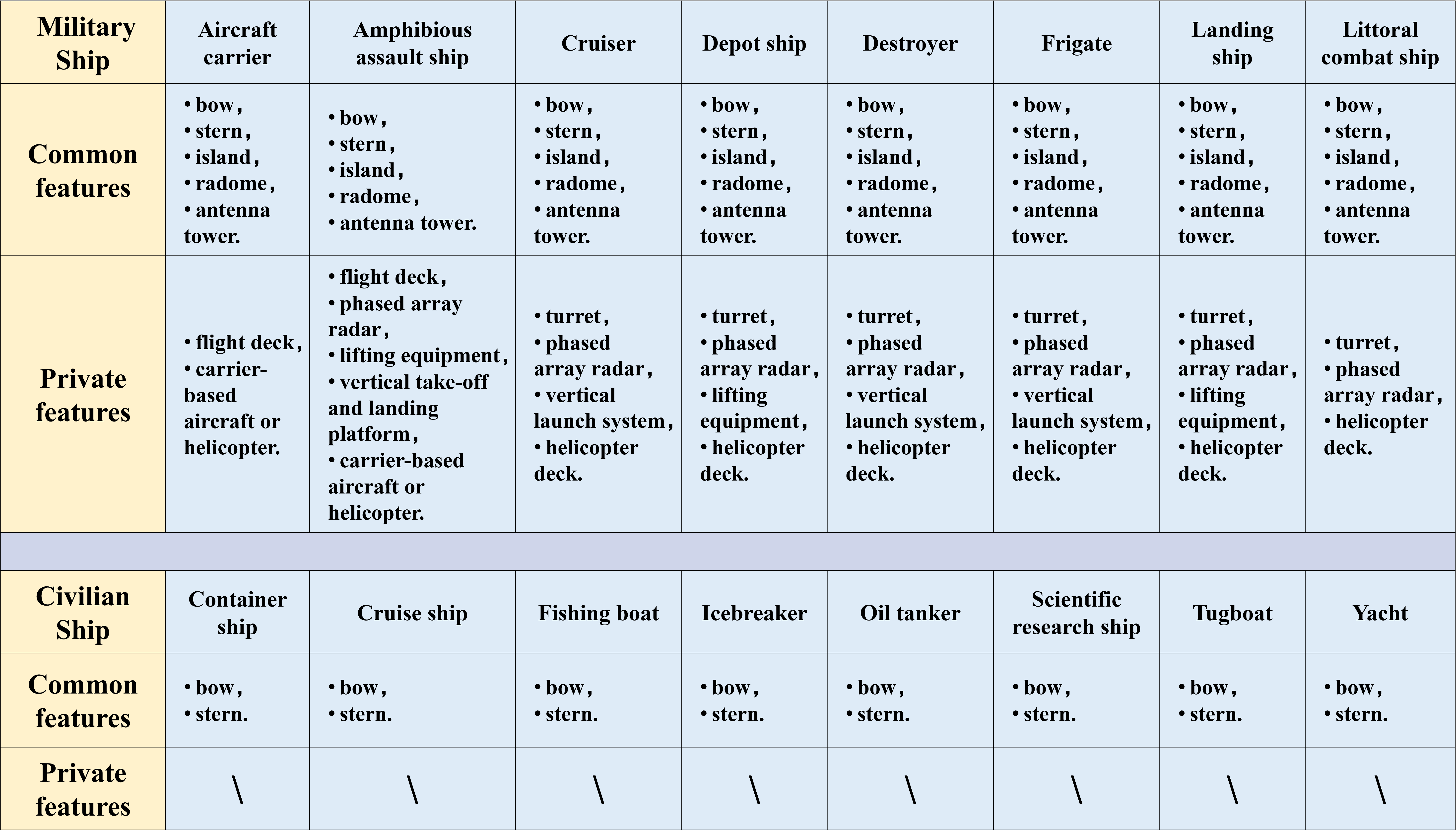}
   \caption{Feature information for 16 fine-grained ship categories in the FGSC domain knowledge base.}
   \label{fig:base}
\end{figure*}

To obtain an instruction-following dataset specific to the FGSC task, we introduce a domain knowledge-enhanced CoT prompt generation mechanism. The CoT prompts designed in this method consist of two parts: “descriptions" and “rules". The “descriptions" part provides domain knowledge-enhanced information for task-specific instruction generation. To ensure that the generated instructions can convey the coupling relationship between imaging conditions and discriminative ship features, we build the domain knowledge base in Step 1 and collect descriptions of both elements from ship images in Step 2 and 3. The “rules" part governs the content and format of the generated instructions. To ensure that the generated instruction data can consider the correct logical relationships among multiple dialogue rounds, we meticulously design the principles of CoT instruction generation in Step 4. Specifically, as shown in \textcolor{blue}{Fig. \ref{fig:method compared} (b)}, our method includes the following four steps:

\indent \textbf{Step 1: Building FGSC domain knowledge base.} To construct a domain knowledge base related to the FGSC task, we first utilize various knowledge search tools (such as Wikipedia, Baidu Baike, Google search engine, etc.) and manually obtain descriptive information about different fine-grained ship types as the source of domain knowledge. However, some of the information contains irrelevant features that cannot be represented in images, such as type of propulsion and displacement. Therefore, we eliminate these irrelevant features by collecting image samples of different fine-grained ship types as references. Further, we categorize the retained ship features into two categories: one consists of common features shared by different ship targets (such as “bow”, “stern”), and the other comprises private features only possessed by certain ship categories (such as “flight deck”, “vertical launch system”). In summary, we develop an FGSC domain knowledge base, as shown in \textcolor{blue}{Fig. \ref{fig:base}}.

\indent \textbf{Step 2: Generating descriptions of imaging conditions in the ship image.} In remote sensing images, both the common and private features of ship targets may become invisible due to the change of imaging conditions. Therefore, we believe that using imaging conditions as prompts can better help VLMs capture the visible features of ship targets. To this end, we design a new task-related prompt to help us obtain descriptions of imaging conditions in ship images. Specifically, the designed prompt is as follows: \textit{“Describe the main information you can see in this picture in a paragraph. The description must include the clarity of the image, the weather in the image, the size of the ship and the influence of these factors on the visible objects on the ship.”} In the designed prompt, the description generation tools are required to describe the imaging conditions of the ship image, such as clarity and weather conditions, as well as the impact of these conditions on the ship target. By inputting the above prompt along with ship images into the description generation tools (such as GPT-4) and making manual corrections, we obtain the descriptions of imaging conditions in the ship image.

\begin{figure*}[ht]
  \centering
   \includegraphics[width=1\linewidth]{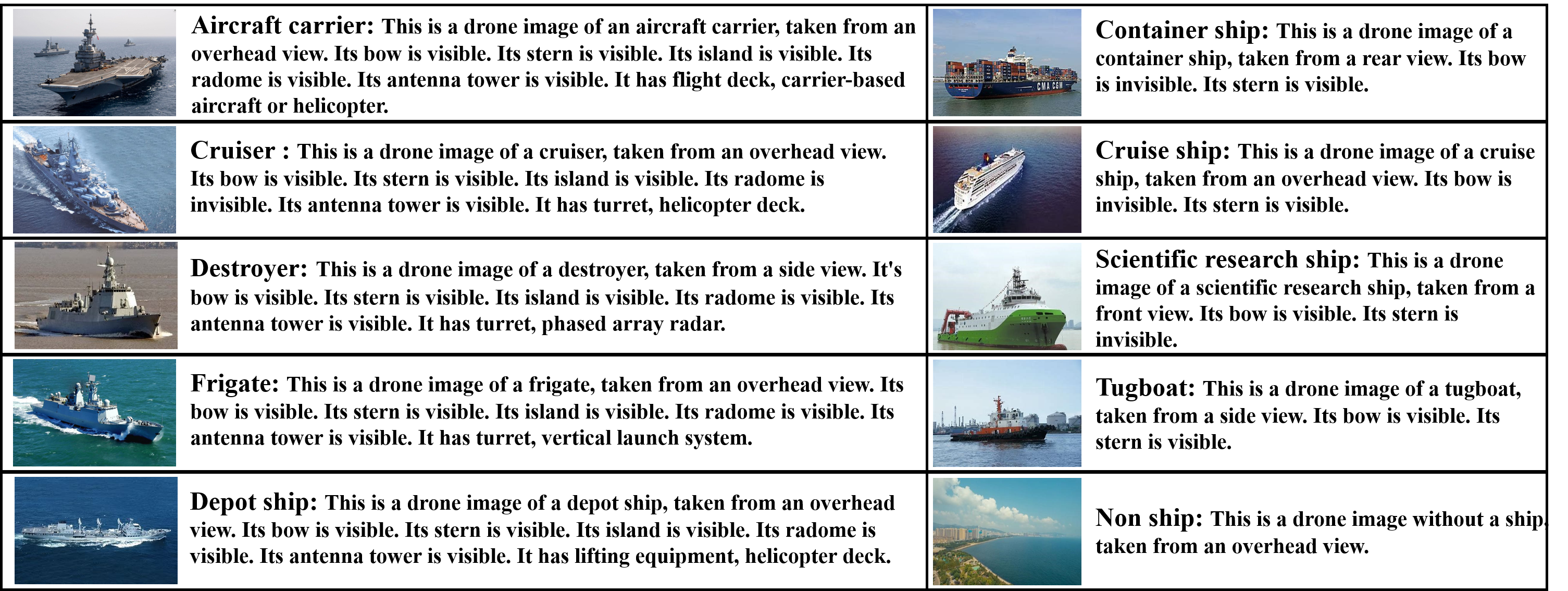}
   \caption{Examples of descriptions of discriminative features for different ship targets.}
   \label{fig:description example}
\end{figure*}

\indent \textbf{Step 3: Generating descriptions of discriminative features for the ship target.} After obtaining the descriptions of imaging conditions, we need to further generate descriptions of visible discriminative features for the ship target. This allows the prompts to accurately express the coupling relationship between imaging conditions and discriminative features. Specifically, we complete this task through manual template filling. As shown in \textcolor{blue}{Fig. \ref{fig:description example}}, based on the FGSC domain knowledge base, the designed templates consist of four main parts: category information, perspective, common features, and private features. For military ships, the description template is as follows: \textit{“This is a drone image of [class], taken from [perspective]. Its bow is [vi/invi]. Its stern is[vi/invi]. Its island is[vi/invi]. Its radome is [vi/invi]. Its antenna tower is [vi/invi]. It has [spart].”} The term \textit{“class”} refers to the fine-grained category of the ship target in the image. \textit{“perspective”} indicates the shooting angle of the ship, \textit{“vi/invi”} requires determining whether features are visible or invisible, and \textit{“spart”} pertains to the ship target's visible private features, as defined in the FGSC domain knowledge base. For civilian ships, type differentiation is straightforward, as their bows and sterns provide distinct identifying features. We also observed that their private features are highly variable, with significant differences even within the same category. Consequently, introducing private features does not notably improve classification, as confirmed by our preliminary experiments. Therefore their description template is simplified as follows: \textit{“This is a drone image of [class], taken from [perspective]. Its bow is [vi/invi]. Its stern is[vi/invi].”} For images without ship targets, a uniform description is applied: \textit{“This is a drone image without a ship, taken from an overhead view.”} Through manual template filling, we obtain descriptions of discriminative features for each ship type under different imaging conditions.

\begin{figure}[h]
  \centering
  \includegraphics[width=0.95\linewidth]{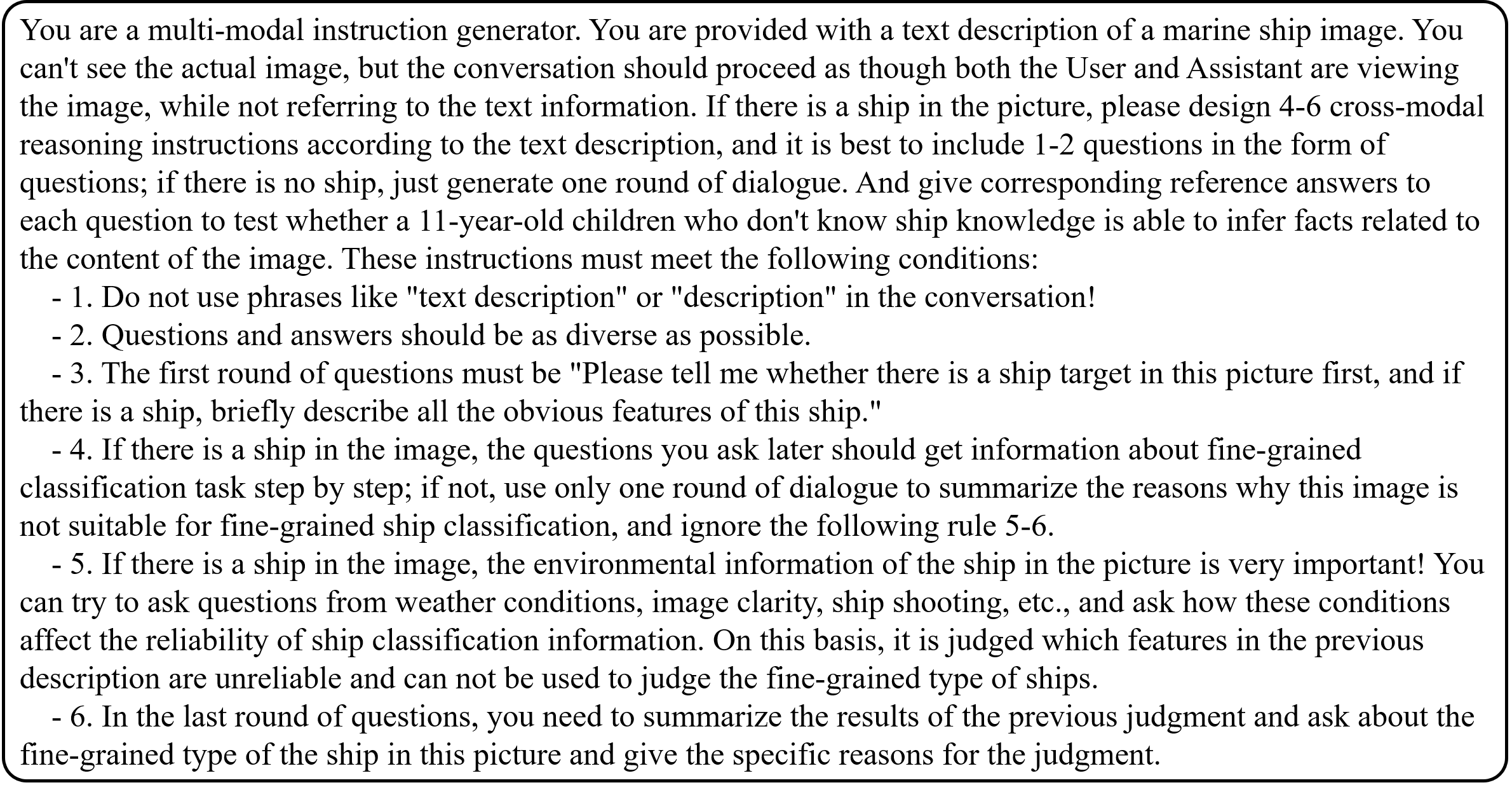}
  \caption{Principles of CoT instruction generation.}
   \label{fig:prompt}
\end{figure}

\begin{figure*}[ht]
  \centering
   \includegraphics[width=0.96\linewidth]{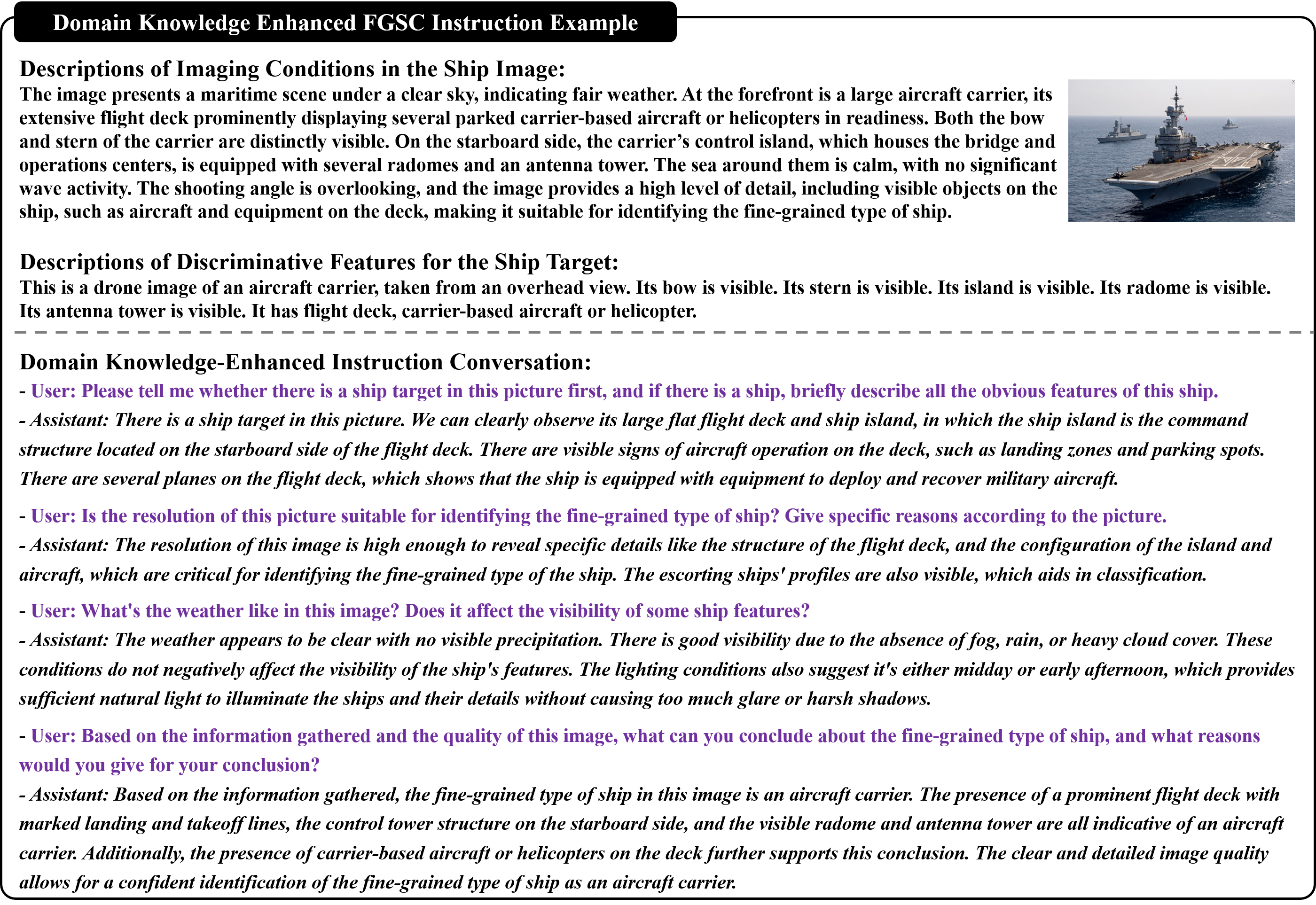}
   \caption{An instruction instance from the TITANIC-FGS dataset.}
   \label{fig:instruction example}
\end{figure*}

\indent \textbf{Step 4: Instruction generation based on domain knowledge-enhanced CoT prompts.} The descriptions obtained from the above three steps ensure that the “descriptions" part of CoT prompts can incorporate detailed information relevant to the FGSC task. In this step, we integrate the principles of CoT instruction generation into the “rules" part to ensure that the generated instructions include task-specific reasoning logic for the FGSC task. Specifically, we redefine the FGSC problem as a step-by-step reasoning problem and design the principles of CoT instruction generation as shown in \textcolor{blue}{Fig. \ref{fig:prompt}}. Among the six principles, the first two are fundamental principles of instructions generation, while the latter four are requirements derived from the reasoning logic chain specific to the FGSC task. Specifically, Principle 3 requires the instructions to first summarize the reliable fine-grained ship features present in the input image. Subsequently, Principle 4 and 5 require instructions to determine how imaging conditions affect the reliability of features. Finally, Principle 6 requires instructions to collect reliable discriminative fine-grained ship features to classify its fine-grained types. The last four principles determine the main logic of multi-round dialogue for FGSC tasks in our CoT instructions. By inputting the CoT prompts into the instruction generation tools (such as GPT-4), we obtain the domain knowledge-enhanced dataset, TITANIC-FGS. For the generated instructions in TITANIC-FGS dataset, we conduct necessary manual review and adjustments. \textcolor{blue}{Fig. \ref{fig:instruction example}} illustrates an instance from the TITANIC-FGS dataset.

Our TITANIC-FGS dataset covers 16 fine-grained ship categories, and 1 category without ship denoted as “Non ship (C9)”. The 16 ship categories include 8 military ships: Aircraft carrier (C1), Amphibious assault ship (C2), Cruiser (C3), Depot ship (C4), Destroyer (C5), Frigate (C6), Landing ship (C7), and Littoral combat ship (C8); and 8 civilian ships: Container ship (C10), Cruise ship (C11), Fishing boat (C12), Icebreaker (C13), Oil tanker (C14), Scientific research ship (C15), Tugboat (C16), and Yacht (C17). In terms of sample number, the TITANIC-FGS dataset comprises 18,929 optical remote sensing images, with 16,876 used for training and the remaining 2,053 for testing. These images are sourced from publicly accessible image search engines, such as Google and Baidu. The images range in size from 200 × 100 to 4000 × 2000 pixels. For each ship image in the training dataset, the instructions consist of 4-6 rounds of dialogue. For images without ship targets, the instructions typically include only one round of dialogue to inquire whether ship targets are present in the image. The testing dataset of TITANIC-FGS only contains the ship images and corresponding fine-grained ship category labels.

To address the manual review tasks in the four implementation steps, we formed an expert group consisting of eight individuals with a strong understanding of ship classification. In the material collection of Step 1, we gathered information from multiple search engines and reference books, while exchanging opinions and comparing with images to ensure the accuracy of the knowledge. Regarding the Step 2, 3 and 4, manual corrections were performed using a division of labor and cross-validation method to ensure the accuracy of descriptions and instructions. For the ship image description generation in Step 2, we utilized the GPT-4 API, which took approximately 28 hours to generate descriptions for all 16,876 images. In Step 3, manual template filling was carried out based on predefined templates, where we populated the features collected in \textcolor{blue}{Fig. \ref{fig:base}} for each type of ship. In Step 4, manual inspection compares the generated instructions with the images to identify incorrect expressions or irrelevant questions, ensuring the logical coherence and accuracy of the multi-round dialogue. The construction of the TITANIC-FGS dataset took approximately 18 days to complete. To the best of our knowledge, the TITANIC-FGS dataset is the first domain knowledge-enhanced instruction-following dataset for the FGSC task.

\subsection{Instruction Tuning Based on TITANIC-FGS for the FGSC Task}
\label{subsec3-3}
\begin{figure}[h]
  \centering
   \includegraphics[width=0.95\linewidth]{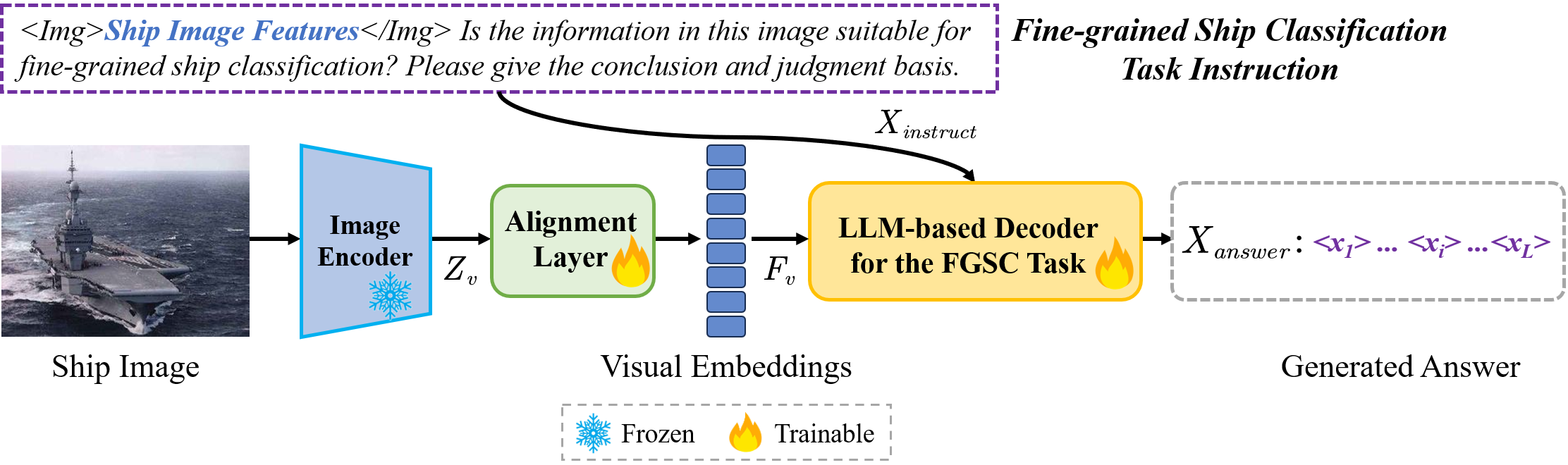}
   \caption{The overall framework of the IFShip model.}
   \label{fig:model}
\end{figure}

After obtaining the TITANIC-FGS dataset, we can begin fine-tuning the IFShip model. As shown in \textcolor{blue}{Fig. \ref{fig:model}}, our IFShip model, built upon the LLaVA, primarily consists of an image encoder, an alignment layer, and an LLM-based decoder for the FGSC task. The specific information is as follows:

\textbf{\textit{Image Encoder.}} The image encoder used in IFShip is the pre-trained CLIP ViT-L model. For an input ship image $I\in{R}^{H \times W \times 3}$, where \textit{H} and \textit{W} respectively represent height and width, its resolution is firstly standardized to 336×336. Then, the image is segmented into multiple image patches of size 14×14. Finally, the image encoder extracts image embeddings ${Z}_{v}\in {R}^{N \times D}$ from these patches, where \textit{N} means the number of image patches and \textit{D} means the hidden dimension of the embeddings.

\textbf{\textit{Alignment Layer.}} The role of the alignment layer is to facilitate the integration of image and text features. It maps the image features ${Z}_{v}$ to the dimensions of the text features, resulting in visual tokens ${F}_{v}=proj({Z}_{v})$, where $proj(\cdot)$ denotes the alignment mapping operation on input features. To enable rapid model iteration, we use an MLP with a structure of two rounds of Linear + GELU + Linear as the alignment layer.

\textbf{\textit{LLM-based Decoder for the FGSC Task.}} The LLM-based decoder receives the instruction ${X}_{instruct}$ related to the FGSC task and the visual tokens ${F}_{v}$ of the ship image. Then it generates the answer ${X}_{answer}$ specific to the instruction requirements. For IFShip, we select Llama-7B as the LLM-based decoder.

In terms of model training, traditional works use a two-stage instruction tuning approach, involving task-specific concept alignment in the first stage and semantic instruction tuning in the second stage. However, we omitted the concept alignment phase during our training. On one hand, the alignment training approach may cause “deceptive alignment” during instruction tuning. On the other hand, existing studies \cite{cite19} show that concept alignment alone does not significantly improve model performance, whereas semantic instruction tuning greatly enhances performance across tasks.

During the instruction tuning process, we keep the weights of the visual encoder frozen while updating the weights of the alignment layer and the LLM-based decoder. We represent the multi-turn dialogue of the instruction data in TITANIC-FGS dataset as a list ${X}_{c}=({X}_{instruct}^{1},{X}_{response}^{1},...,{X}_{instruct}^{n},{X}_{response}^{n})$, where ${X}_{instruct}^{n}$ denotes the instruction for the ${n}_{th}$ turn and ${X}_{response}^{n}$ represents the corresponding response. Therefore, the objective function ${f}_{target}$ is defined as follows:

\begin{equation}
\begin{aligned}
{f}_{target} & = logP({X}_{response}|F_{v},{X}_{instruct};\theta) \\
& = \displaystyle\sum_{i=1}^{L}logP({x}_{i}|F_{v},{X}_{instruct,<i},{X}_{response,<i};\theta)
\end{aligned}
\end{equation}
where ${F}_{v}$ represents the ship image feature, $P$ and $\theta$ are the conditional probability and the trainable parameters of the model, respectively. ${X}_{instruct,<i}$ and ${X}_{response,<i}$ respectively denote the instruction tokens and response tokens before the current ${i}_{th}$ predicted response, $L$ denotes the length of ${X}_{response}$. In this stage, we use the Low-Rank Adaptation (LoRA) method to fine-tune IFShip’s alignment layer and LLM-based decoder.

\subsection{FGSC Visual Chatbot}
\label{subsec3-4}
\begin{figure*}[ht]
  \centering
   \includegraphics[width=1\linewidth]{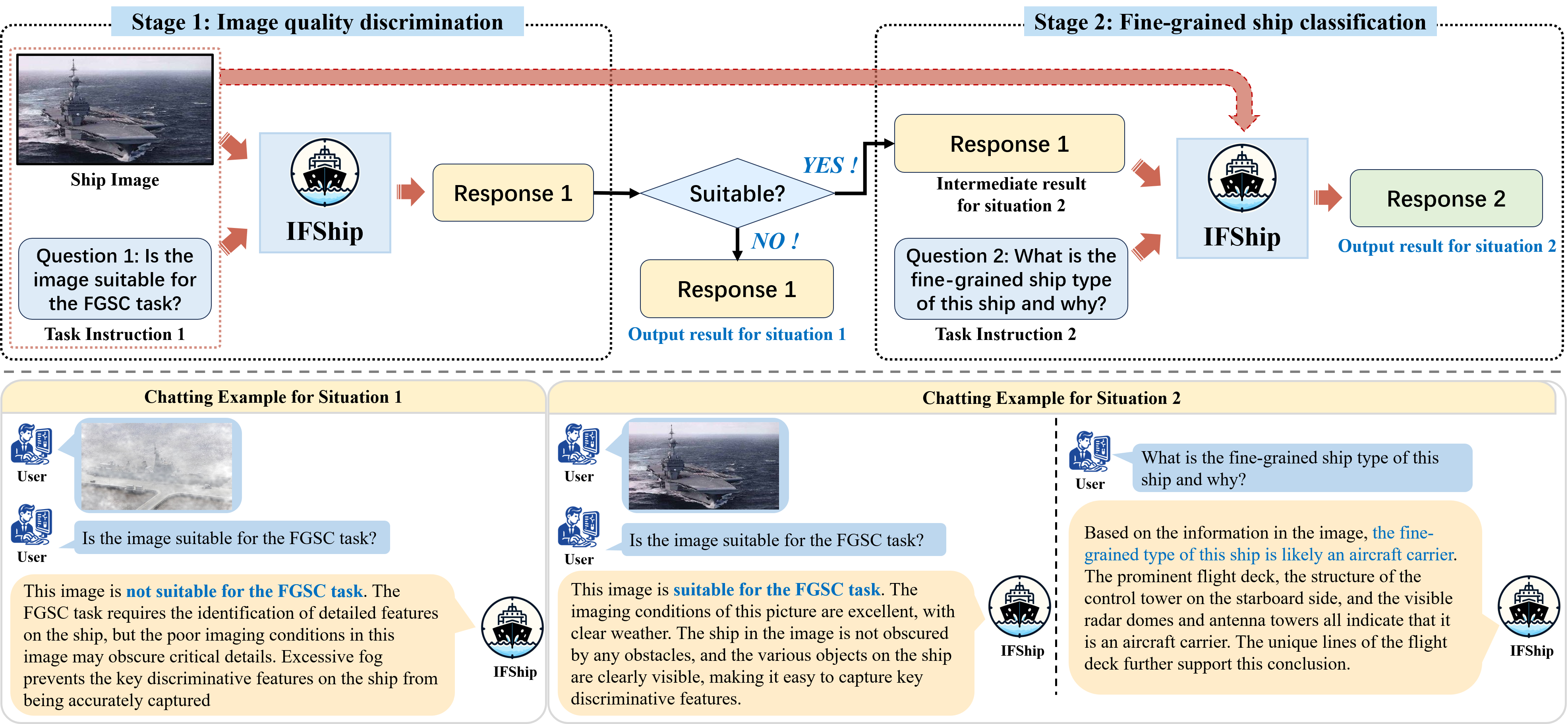}
   \caption{Top: The FGSC visual chatbot based on IFShip. Bottom: Two examples based on the chatbot.}
   \label{fig:chatbot}
\end{figure*}

Building on IFShip, we design an interpretable FGSC visual chatbot. To address the limitations of traditional end-to-end FGSC algorithms, which lack interpretability and often misclassify images, we propose a two-stage classification paradigm, as shown in \textcolor{blue}{Fig. \ref{fig:chatbot}}. In our classification paradigm, the FGSC visual chatbot first determines whether the input image is suitable for the FGSC task, thereby eliminating the impact of unsuitable images. For ship images with sufficient discriminative features, the FGSC visual chatbot then proceeds to the second stage to give the correct fine-grained ship category of the target.

Specifically, the FGSC visual chatbot first combines the given ship image with the task instruction 1 (\textit{“Is the image suitable for the FGSC task?”}) and feed them into the IFShip to determine whether the image is suitable for FGSC. If Response 1 indicates that the image is unsuitable, the FGSC visual chatbot will use Response 1 as the output for this task and will not proceed to the stage 2. Otherwise, if Response 1 indicates that the image is suitable, the FGSC visual chatbot will proceed to the stage 2. At this stage, the FGSC visual chatbot inputs the Task instruction 1, Response 1, the ship image, and the designed Task instruction 2 (\textit{“What is the fine-grained ship type of this ship and why?”}) into IFShip, generating Response 2 that includes the FGSC result and the reasons for the classification. This allows the model to understand the task context better by considering the complete dialogue history. For this situation, Response 2 becomes the output for the task. As shown in \textcolor{blue}{Fig. \ref{fig:chatbot}}, the FGSC visual chatbot can handle different situations with the two-stage classification paradigm. In situation 1 where the image quality is unsuitable for the FGSC task, it provides a reason as the final task output. This approach avoids unreliable classifications without accurate justification. In situation 2 where the image quality is suitable, the FGSC visual chatbot conducts second round of dialogue and provides the classification results along with reliable reasons for the classification.

To evaluate the interpretability of the model's classification results, we utilized GPT-4 to construct two testing datasets based on the testing image samples of TITANIC-FGS dataset: the Fine-grained Ship Image Caption Dataset and the Fine-grained Ship Image VQA Dataset. The reason we design these two tasks is that we believe the interpretability of ship classification results essentially equates to the model’s ability to accurately capture and express the key information in ship images.

\begin{figure*}[ht]
  \centering
   \includegraphics[width=0.95\linewidth]{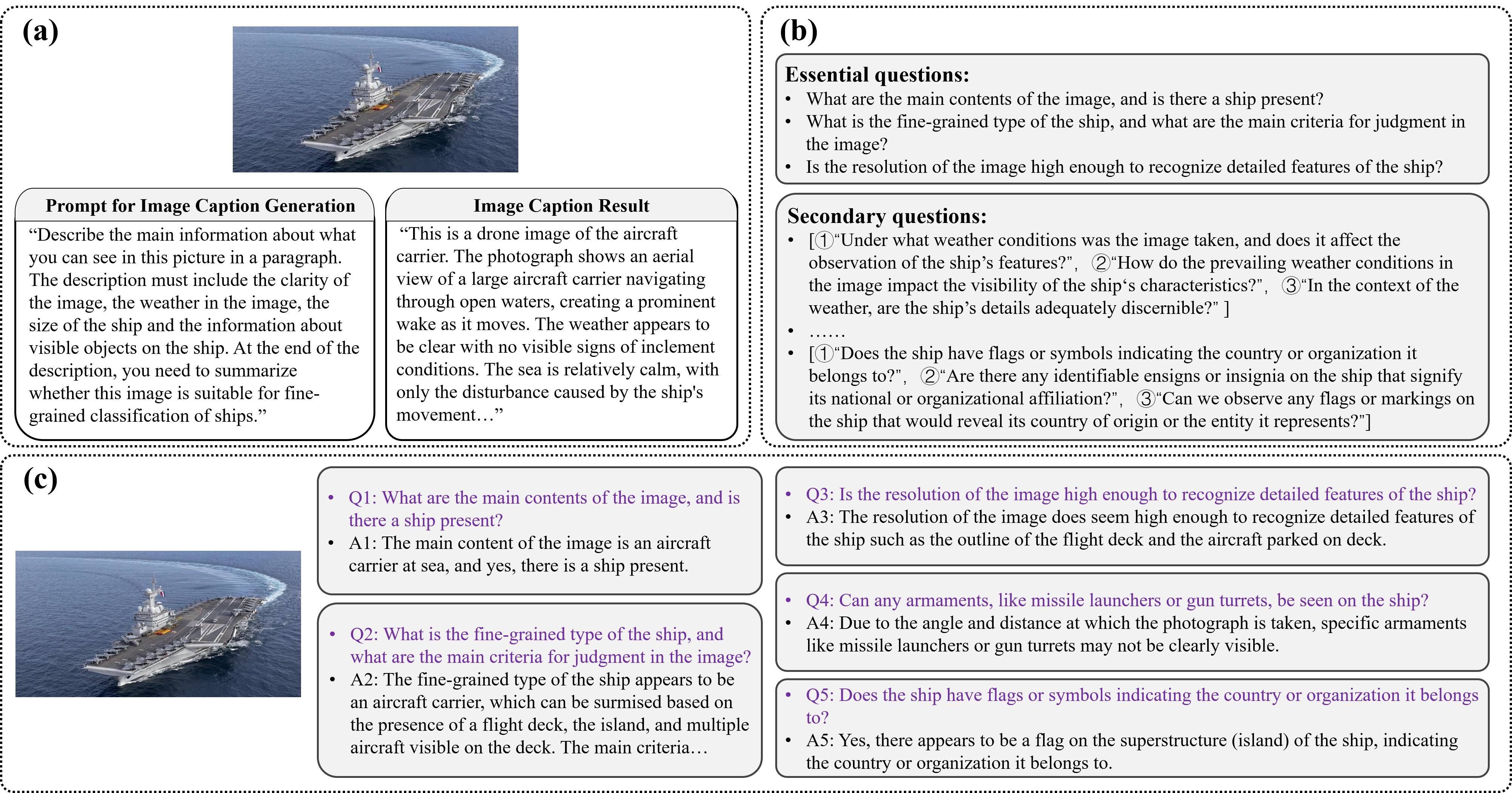}
   \caption{(a) Caption generation prompt and the result sample of the Fine-grained Ship Image Caption dataset. (b) Examples of question library used to construct the Fine-grained Ship Image VQA datasets. (c) The testing sample of the Fine-grained Ship Image VQA dataset.}
   \label{fig:test dataset}
\end{figure*}

\textbf{\textit{Fine-grained Ship Image Caption Dataset.}} This dataset aims to test the model's ability to understand ship image information, and requires the model to generate descriptions of the content in ship images. In the dataset construction process, we first design a caption generation prompt, as shown in \textcolor{blue}{Fig. \ref{fig:test dataset} (a)}. The prompt is combined with each image and input into the GPT-4 model to generates a corresponding caption for the image. The generated captions are then manually reviewed to remove irrelevant content and correct any inaccuracies based on the image content. Using 2,053 test images from the TITANIC-FGS dataset, we generate a total of 2,053 caption samples. \textcolor{blue}{Fig. \ref{fig:test dataset} (a)} illustrates an example of the image caption task.

\textbf{\textit{Fine-grained Ship Image VQA Dataset.}} This dataset evaluates the model's ability to extract ship image information based on task instructions. To construct this dataset, we first create a unified question library comprising two categories: essential questions and secondary questions, as shown in \textcolor{blue}{Fig. \ref{fig:test dataset} (b)}. Essential questions focus on fundamental details of the image, such as its content, ship category, and image quality, while secondary questions address additional descriptive aspects, including weather conditions and ship movement. The library contains three types of essential questions and nine types of secondary questions, each secondary question having three variations. When a secondary question is selected, one of its variations is randomly chosen. For images containing ship targets, GPT-4 is required to answer all three essential questions and a random subset of secondary questions. For images without ship targets, we only ask questions about the content and the presence of ships. The generated answers are then manually reviewed to remove irrelevant content and correct any inaccuracies based on the image content. Using 2,053 test images from the TITANIC-FGS dataset, we construct 11,005 image VQA samples, as the number of questions varies across images. \textcolor{blue}{Fig. \ref{fig:test dataset} (c)} illustrates an example of the resulting VQA dataset.

\section{Experiments and Results}
\label{subsec4}
\subsection{Experimental Details}
\label{subsec4-1}
In this section, we conduct experiments on three tasks: fine-grained ship classification, ship image caption and ship image VQA. The experimental details are as follows:

\textbf{\textit{1) Dataset:}} For the FGSC task, we first evaluate IFShip and SOTA FGSC methods on the TITANIC-FGS dataset. \textcolor{blue}{Table \ref{tab:num}} illustrates the distribution of image samples in the TITANIC-FGS dataset. To evaluate generalization, we also introduce the publicly available FGSCR42 dataset \cite{cite14}, construct a test dataset based on categories aligned with our classification system. This dataset contains 4,534 satellite ship images, with detailed information provided in \textcolor{blue}{Table \ref{tab:num2}}. For the fine-grained ship image caption task and the ship image VQA task, we evaluate IFShip and SOTA VLMs on the Fine-grained Ship Image Caption Dataset and Fine-grained Ship Image VQA Dataset, which are introduced in \textcolor{blue}{Subsection \ref{subsec3-4}}.

\textbf{\textit{2) Implementation Details:}} For IFShip, we conduct training for only 1 epoch on the TITANIC-FGS dataset. The AdamW is used as the optimizer, and we set the batch size to 128 with ${2 \times {10}^{-4}}$ learning rate. The training is conducted on one NVIDIA A6000 GPU (48 GB memory), completing the task in 6 hours. The rank for LoRA is set to 128.

\textbf{\textit{3) Comparative Methods:}} For the FGSC task, we selected a set of representative and SOTA fine-grained classification methods. These include CNN-based approaches, such as BCNN(VGG-16,VGG-19) \cite{cite52}, DCL(ResNet-50) \cite{cite53}, TASN(ResNet-50) \cite{cite54}, VAN(VAN-B0) \cite{cite55}, and {P\textsuperscript{2}Net}(ResNet-50) \cite{cite1}, as well as Transformer-based methods, including ViT-L \cite{cite51}, ACC-ViT(ViT-B) \cite{cite71}, and IELT(ViT-B) \cite{cite70}. Additionally, we introduced the classic VLM with both a visual encoder and a text encoder, CLIP (ViT-L) \cite{cite16}. For these comparative methods, we train them on the image dataset of TITANIC-FGS. The batch size is set to 16, and the number of training epochs is set to 100. To ensure fair comparison, the input image size for these methods is standardized to 336×336, consistent with IFShip's image size for data augmentation.

For the ship image caption and ship image VQA tasks, the comparative VLMs include LLaVA(Llama-7B) \cite{cite18}, MiniGPT-4 \cite{cite24}, MoE-LLaVA \cite{cite25}, Bunny(Phi2-2.7B) \cite{cite26}, InstructBLIP(Vicuna0-7B) \cite{cite27} and GeoChat(Vicuna1.5-7B) \cite{cite69}. We include two variants with different backbones of MiniGPT-4(Llama2-7B/Vicuna0-13B) and MoE-LLaVA(Phi2-2.7B/Qwen-2.2B). These models are representative and advanced open-source VLMs in the general domain and RS domain. These VLMs are not fine-tuned but only loaded with the pre-trained model weights from the general dataset to test their zero-shot capabilities on the two tasks.

\begin{table}[ht]
  \centering
  \Large
  \caption{Distribution of the TITANIC-FGS dataset.}
  \renewcommand{\arraystretch}{1} 
  \resizebox{1\columnwidth}{!}{
    \begin{tabular}{c|*{17}{c}}
    \hline
    \multirow{2}{*}{} & \multicolumn{17}{c}{Ship Class} \\ \cline{2-18} 
                  & C1 & C2 & C3 & C4 & C5 & C6 & C7 & C8 & C9 & C10 & C11 & C12 & C13 & C14 & C15 & C16 & C17 \\ \hline
    Train         & 1780 & 1575 & 1218 & 677 & 3297 & 1962 & 459 & 1011 & 700 & 435 & 817 & 287 & 392 & 793 & 322 & 418 & 733 \\ \hline
    Test          & 138 & 167 & 113 & 98 & 169 & 132 & 122 & 103 & 111 & 122 & 126 & 109 & 103 & 120 & 100 & 114 & 106 \\ \hline
    \end{tabular}}
    \label{tab:num}
\end{table}

\begin{table}[ht]
  \centering
  \Large
  \caption{Distribution of the FGSCR42 dataset.}
  \renewcommand{\arraystretch}{1} 
  \resizebox{0.5\columnwidth}{!}{
    \begin{tabular}{c|*{7}{c}}
    \hline
    \multirow{1}{*}{} & \multicolumn{7}{c}{Ship Class} \\ \cline{2-8} 
                  & C1 & C3 & C5 & C6 & C7 & C10 & C17 \\ \hline
    Test          & 933 & 607 & 1094 & 88 & 394 & 455 & 963 \\ \hline
    \end{tabular}}
    \label{tab:num2}
\end{table}

\subsection{Evaluation Metrics}
\label{subsec4-2}
For FGSC task, we use the accuracy rate (AR) for each category and the overall accuracy (OA) as evaluation metrics. Additionally, we develop an evaluation metric for classification reliability named reliability score (${S}_{rel}$). It is defined as follows:
\begin{equation}
\begin{aligned}
{S}_{rel} = {p}_{conf}log({p}_{conf})
\label{eq:shi1}
\end{aligned}
\end{equation}
where ${p}_{conf}$ represents the model's confidence level. For end-to-end FGSC methods and CLIP, the highest category probability is used as their confidence level. For IFShip, the confidence level is set to 1 if the model deems an image suitable for FGSC and 0 if unsuitable. To ensure that the ${S}_{rel}$ increases monotonically and the value range of ${S}_{rel}$ is [0,1], we optimize the \textcolor{blue}{Eq. \ref{eq:shi1}}. Furthermore, we incorporate the classification accuracy ${p}_{acc}$ into the metric, allowing it to reward or penalize based on the correctness of the classification results. In summary, the ${S}_{rel}$ is optimized as follows:
\begin{equation}
\begin{aligned}
{S}_{rel} = \frac{{p}_{conf}log(\frac{{p}_{conf}}{\gamma }+1)}{log(\frac{1}{\gamma }+1)}\cdot {p}_{acc}
\label{eq:shi2}
\end{aligned}
\end{equation}
where the confidence threshold ${\gamma}$ is a constant that ranges from 0 to 1. For the classification accuracy ${p}_{acc}$, if the classification result is correct, it is 1; otherwise, it is -1. In the experiments, we set ${\gamma}$ to 0.6. Based on \textcolor{blue}{Eq. \ref{eq:shi2}}, we calculate the reliability score rate (SR) for each ship category and the overall average reliability score (OS) as an additional evaluation metric for FGSC task.

For the image caption task and image VQA task, we employ a LVLM-based evaluation method. By designing specific prompts, we ask the expert LVLM to determine whether the generated results cover all the visual information and correct answers in the ground truth. If the expert LVLM returns “Yes”, the result is considered correct; otherwise, it is considered incorrect. Finally, the overall accuracy (OA) of each VLM on all testing samples is calculated as the performance metric of the model. During the experiments, we select two expert LVLMs: GPT-3.5 and GPT-4.0.

\subsection{Results of Accuracy and Reliability for the FGSC Task}
\label{subsec4-3}
In the paradigm we previously designed, IFShip only classifies images it deems suitable. To make the comparison more comprehensive and fair, we tested two scenarios: IFShip(jud), which classifies only suitable samples, and IFShip(all), which classifies all images without the judgment stage. The AR of each category and OA of the TITANIC-FGS testing dataset are shown in \textcolor{blue}{Table \ref{tab:class1}}; the SR of each category and OS are shown in \textcolor{blue}{Table \ref{tab:rel1}}. For FGSCR42 dataset, since IFShip considers all images suitable for classification, the experimental results are uniformly presented as “IFShip”. All the experimental results of the FGSCR42 dataset are shown in \textcolor{blue}{Table \ref{tab:new}}.

\textbf{\textit{Accuracy of Fine-grained Ship Classification Results:}} From the results presented in \textcolor{blue}{Table \ref{tab:class1}}, it can be observed that IFShip(jud) exhibits significant performance advantages compared to other FGSC methods and CLIP. Regarding the OA, IFShip(jud) outperforms other methods with a performance of 94.02\%. Compared to the model with the highest accuracy in end-to-end algorithms, ACC-ViT, IFShip(jud) exhibits accuracy improvements of 11.75\%. Even for IFShip(all), which classifies all testing images, the OA reaches 85.44\%, outperforming all the comparative methods. For the AR for each ship category, IFShip(jud) demonstrates an absolute lead across the majority of classes, achieving 100\% accuracy in the C7, C9, C10, C14, and C15. IFShip(all) also achieves accuracy second only to IFShip(jud) in C7, C9, and C14. For the most easily confused categories: C3 (Cruiser), C5 (Destroyer), and C6 (Frigate), the comparative methods exhibit significant performance deficiencies. The appearance of these three types of ships is highly similar, and distinguishing them requires relying on key local features. However, for C3 and C5, the highest accuracy among the comparative methods is achieved by CLIP, with accuracies of 65.41\% and 60.71\%, respectively. Although {P\textsuperscript{2}Net} achieves an excellent accuracy of 86.87\% in C6, the accuracies of other comparative methods do not exceed 70\%. In contrast, IFShip(jud) exhibits improvements of 23.72\%, 30.49\%, and 3.23\% in C3, C5 and C6, respectively. IFShip(all) also achieves improvements of 15.29\% and 17.73\% in C3 and C5.

From the results in \textcolor{blue}{Table \ref{tab:new}}, all classification models show poor overall accuracy. While IFShip performs well on the seven subcategories, its overall accuracy is only 67.03\%, as it may misclassify samples into non-existing categories. Similar issues are observed in other comparative methods. Although IFShip's OA is not very high, it still outperforms other methods, showing a significant 26.60\% advantage over the second-best ACC-ViT. While the comparative methods perform well only on easily classified categories C10(Container ship) and C17(Yacht), IFShip maintains stable performance across all categories.

\begin{table*}[htbp]
  \centering
  \huge
  \caption{Accuracy comparative results of different methods on TITANIC-FGS dataset.}
  \renewcommand{\arraystretch}{1} 
  \resizebox{1\columnwidth}{!}{
    \begin{tabular}{l|*{17}{c}|c}
    \hline
    \multirow{2}{*}{\textbf{Method}} & \multicolumn{17}{c|}{\textbf{AR}} & \multirow{2}{*}{\textbf{OA}} \\ \cline{2-18}
                         & \textbf{C1} & \textbf{C2} & \textbf{C3} & \textbf{C4} & \textbf{C5} & \textbf{C6} & \textbf{C7} & \textbf{C8} & \textbf{C9} & \textbf{C10} & \textbf{C11} & \textbf{C12} & \textbf{C13} & \textbf{C14} & \textbf{C15} & \textbf{C16} & \textbf{C17} \\ \hline
    BCNN\cite{cite52}                 & 71.02 & 84.38 & 51.41 & 84.55 & 37.34 & 51.52 & 80.85 & 88.57 & 92.04 & 88.98 & 90.98 & 79.28 & 78.82 & 81.03 & 94.59 & 84.71 & 89.81 & 75.84               \\
    DCL\cite{cite53}                  & 75.86 & 77.65 & 50.34 & 91.40 & 37.84 & 48.10 & 61.19 & 69.92 & 80.92 & 85.25 & 91.20 & 78.82 & 80.56 & 82.11 & 89.71 & 91.67 & 63.92 & 71.21               \\
    TASN\cite{cite54}                 & 77.30 & 82.10 & 51.91 & 83.18 & 29.13 & 49.23 & 80.77 & 96.63 & 93.86 & 88.28 & 89.92 & 83.00 & 84.88 & 88.30 & 87.80 & 90.79 & 78.46 & 74.09               \\
    VAN\cite{cite55}                  & 81.10 & 88.37 & 60.47 & \textbf{94.95} & 41.07 & 52.00 & 83.18 & 93.27 & 94.69 & 93.22 & 89.78 & 85.29 & 77.66 & 88.18 & 97.40 & 90.42 & 90.45 & 80.42               \\
    {P\textsuperscript{2}Net}\cite{cite1}                & 71.98 & 85.71 & 55.71 & 83.33 & 36.96 & 86.87 & 83.51 & 92.00 & 90.00 & 93.50 & 95.97 & 86.87 & 80.43 & 91.09 & 89.74 & 93.26 & 86.73 & 78.13               \\
    ViT-L\cite{cite51}             & 82.39 & 87.50 & 59.20 & 87.96 & 42.45 & 50.30 & 73.50 & 91.67 & \textbf{100.00} & 97.41 & 95.42 & 91.43 & 88.76 & 88.79 & 82.61 & 89.52 & 92.86 & 81.00               \\
    ACC-ViT\cite{cite71}             & 81.48 & 85.88 & 58.46 & 94.00 & 42.00 & 57.64 & 80.37 & 93.94 & 97.37 & 94.44 & 97.62 & 83.19 & 86.41 & 92.24 & 86.90 & 94.90 & 97.14 & 82.27               \\
    IELT\cite{cite70} & 69.63 & 87.25 & 55.47 & 92.47 & 36.11 & 51.18 & 81.32 & 88.46 & 93.91 & 95.76 & 93.89 & 83.62 & 81.90 & 91.26 & 92.21 & \textbf{96.25} & 94.50 & 77.98               \\
    CLIP\cite{cite16} & 79.39 & 79.75 & 65.41 & 79.21 & 60.71 & 70.00 & 67.35 & \textbf{97.73} & 72.01 & 94.45 & \textbf{98.39} & \textbf{95.51} & 90.82 & 87.50 & 81.11 & 94.95 & 97.17 &  81.59               \\ \hline
    IFShip(all) & 71.10 & 75.79 & 80.70 & 84.85 & 78.44 & 85.84 & \textbf{100.00} & 79.13 & \textbf{100.00} & 94.26 & 86.81 & 89.29 & 95.45 & 96.19 & 94.12 & 76.98 & 93.58 & 85.44               \\
    IFShip(jud)               & \textbf{97.25} & \textbf{95.19} & \textbf{89.13} & 86.21 & \textbf{91.20} & \textbf{90.10} & \textbf{100.00} & 89.69 & \textbf{100.00} & \textbf{100.00} & 95.45 & 88.54 & \textbf{97.50} & \textbf{100.00} & \textbf{100.00} & 83.64 & \textbf{98.04} & \textbf{94.02}               \\ \hline
    \end{tabular}}
    \label{tab:class1}
\end{table*}

\begin{table*}[ht]
  \centering
  \huge
  \caption{Reliability comparative results of different methods on TITANIC-FGS dataset(${\gamma}$=0.6).}
  \renewcommand{\arraystretch}{1} 
  \resizebox{1\columnwidth}{!}{
    \begin{tabular}{l|*{17}{c}|c}
    \hline
    \multirow{2}{*}{\textbf{Method}} & \multicolumn{17}{c|}{\textbf{SR}} & \multirow{2}{*}{\textbf{OS}} \\ \cline{2-18}
                         & \textbf{C1} & \textbf{C2} & \textbf{C3} & \textbf{C4} & \textbf{C5} & \textbf{C6} & \textbf{C7} & \textbf{C8} & \textbf{C9} & \textbf{C10} & \textbf{C11} & \textbf{C12} & \textbf{C13} & \textbf{C14} & \textbf{C15} & \textbf{C16} & \textbf{C17} \\ \hline
    BCNN\cite{cite52}     & 61.48 & 44.54 & 17.05 & 70.18 & -30.75 & 14.07 & 19.51 & 58.13 & 61.25 & 52.00 & 65.66 & 36.64 & 25.61 & 39.36 & 25.22 & 23.33 & 63.06 & 36.23 \\
    DCL\cite{cite53}      & 45.88 & 52.41 & 20.05 & 54.67 & -32.50 & 8.81  & 30.78 & 51.87 & 72.04 & 56.42 & 62.37 & 18.84 & 10.07 & 23.91 & 18.81 & 10.52 & 67.33 & 32.34 \\
    TASN\cite{cite54}     & 67.80 & 51.58 & 12.77 & 63.89 & -17.42 & -6.91 & 11.00 & 60.98 & 78.09 & 74.67 & 75.12 & 46.61 & 36.66 & 41.73 & 37.10 & 22.71 & 83.28 & 42.21 \\
    VAN\cite{cite55}      & 55.37 & 48.87 & 20.39 & 57.28 & -11.24 & 10.77 & 31.89 & 54.29 & 62.13 & 46.53 & 58.24 & 34.32 & 28.32 & 41.03 & 33.08 & 30.67 & 59.43 & 37.66 \\
    {P\textsuperscript{2}Net}\cite{cite1}    & 72.44 & 49.90 & 28.40 & 70.18 & -24.45 & 23.52 & 30.82 & 72.07 & 76.21 & 75.41 & 81.70 & 50.92 & 46.09 & 57.49 & 42.10 & 41.83 & 76.57 & 49.53 \\
    ViT-L\cite{cite51}      & 71.55 & 58.17 & 24.73 & 72.03 & -23.02 & 31.38 & 31.90 & 56.46 & 76.87 & 67.46 & 75.64 & 59.33 & 45.99 & 55.93 & 36.15 & 52.27 & 75.11 & 48.62 \\
    ACC-ViT\cite{cite71}      & 63.01 & 51.02 & 19.20 & 64.33 & -28.40 & 10.18 & 25.83 & 53.93 & 67.71 & 63.92 & 67.56 & 56.03 & 49.29 & 53.04 & 31.18 & 43.57 & 65.12 & 42.52 \\
    IELT\cite{cite70}      & \textbf{83.94} & 49.11 & 17.49 & 70.94 & -16.09 & -6.53 & 20.29 & 70.83 & 86.54 & 76.56 & 87.11 & 69.02 & 62.02 & 51.04 & 36.76 & 33.37 & 85.90 & 49.41 \\
    CLIP\cite{cite16} & 75.05 & 35.25 & 23.00 & 43.13 & 0.50 & 9.52 & -6.00 & 40.87 & \textbf{88.04} & 80.77 & 75.85 & 35.48 & 56.79 & 55.97 & 18.52 & 43.20 & 89.11 &  43.70               \\ \hline
    IFShip(all) & 78.26 & \textbf{72.46} & 62.83 & 71.43 & 55.03 & 46.97 & 22.95 & 76.70 & 85.59 & \textbf{88.52} & \textbf{98.41} & \textbf{83.49} & 63.11 & 68.33 & \textbf{60.00} & \textbf{87.72} & \textbf{92.45} & 70.87               \\
    IFShip(jud)   & 75.36 & 57.49 & \textbf{69.03} & \textbf{74.49} & \textbf{60.36}  & \textbf{61.36} & \textbf{46.72} & \textbf{83.50} & 85.59 & 79.51 & 83.33 & 75.23 & \textbf{66.99} & \textbf{71.67} & \textbf{60.00} & 78.07 & 91.51 & \textbf{70.97} \\ \hline
    \end{tabular}}
    \label{tab:rel1}
\end{table*}

\begin{table*}[htbp]
  \centering
  \huge
  \caption{Accuracy and Reliability comparative results of different methods on FGSCR42 dataset(${\gamma}$=0.6).}
  \renewcommand{\arraystretch}{1} 
  \resizebox{1\columnwidth}{!}{
    \begin{tabular}{l|*{7}{c}|*{7}{c}|c|c}
    \hline
    \multirow{2}{*}{\textbf{Method}} & \multicolumn{7}{c|}{\textbf{AR}} & \multicolumn{7}{c|}{\textbf{SR}} & \multirow{2}{*}{\textbf{OA}} & \multirow{2}{*}{\textbf{OS}} \\ \cline{2-15}
                         & \textbf{C1} & \textbf{C3} & \textbf{C5} & \textbf{C6} & \textbf{C7} & \textbf{C10} & \textbf{C17} & \textbf{C1} & \textbf{C3} & \textbf{C5} & \textbf{C6} & \textbf{C7} & \textbf{C10} & \textbf{C17} & \\ \hline
    BCNN\cite{cite52}                 & 42.87 & 9.33 & 38.10 & 2.13 & 0.00 & 96.86 & 94.06 & 33.11 & -33.66 & -33.53 & -42.74 & -27.91 & 22.37 & -3.72 & 34.89 & -7.58               \\
    DCL\cite{cite53}                  & 39.15 & 20.06 & 16.83 & 11.11 & 0.00 & 95.92 & 87.68 & 12.41 & -42.78 & -52.38 & -54.16 & -61.54 & -40.75 & 10.35 & 28.67 & -24.10              \\
    TASN\cite{cite54}                 & 52.56 & 14.29 & 20.85 & 5.29 & 0.00 & 97.92 & 61.41 & 16.76 & -46.22 & -45.43 & -46.95 & -48.75 & -2.10 & 3.06 & 30.02 & -18.41              \\
    VAN\cite{cite55}                  & 40.11 & 12.90 & 34.00 & 0.00 & 14.29 & 94.62 & 92.33 & 11.43 & -25.01 & -19.76 & -39.23 & -22.84 & -1.86 & -5.38 & 25.87 & -9.84              \\
    {P\textsuperscript{2}Net}\cite{cite1}                & 34.63 & 0.00 & 29.79 & 0.00 & 0.00 & 97.56 & 92.56 & 42.43 & -75.98 & -57.05 & -77.29 & -74.99 & -6.76 & 7.37 & 35.36 & -22.34              \\
    ViT-L\cite{cite51}             & 44.52 & 40.00 & 53.45 & 0.00 & 21.05 & 99.22 & 99.80 & 58.01 & -74.74 & -59.41 & -85.86 & \textbf{-0.73} & 20.57 & 5.26 & 36.66 & -17.22               \\
    ACC-ViT\cite{cite71}             & 39.72 & 44.44 & 69.84 & 0.00 & 37.50 & 99.36 & 99.42 & 75.54 & -78.92 & -64.13 & -85.64 & -78.12 & 34.83 & 4.80 & 40.43 & -14.43              \\
    IELT\cite{cite70}         & 40.30 & 5.75 & 33.63 & 0.00 & 0.00 & \textbf{100.00} & 99.03 & 77.25 & -82.63 & -53.77 & -90.04 & -77.13 & -34.20 & -11.43 & 36.19 & -22.45             \\
    CLIP\cite{cite16}         & 30.77 & 22.41 & 12.50 & 0.00 & 0.00 & 96.22 & 98.10 & 79.56 & -54.95 & -59.75 & -59.04 & -60.44 & 19.69 & 28.74 & 38.73 & -3.72             \\ \hline
    IFShip               & \textbf{80.67} & \textbf{93.26} & \textbf{91.99} & \textbf{100.00} & \textbf{100.00} & 98.48 & \textbf{99.84} & \textbf{95.93} & \textbf{9.39} & \textbf{4.94} & \textbf{54.55} & -69.54 & \textbf{99.12} & \textbf{32.50} & \textbf{67.03} & \textbf{34.05}             \\ \hline
    \end{tabular}}
    \label{tab:new}
\end{table*}

\textbf{\textit{Reliability of Fine-grained Ship Classification Results:}} In \textcolor{blue}{Table \ref{tab:rel1}}, the IFShip(judge) achieves a leading OS of 70.97\%, representing an improvement of 21.44\% over the second-best method, P\textsuperscript{2}Net. IFShip(all) also achieves an excellent OS of 70.87\%. For each subcategory, IFShip(jud) and IFShip(all) generally achieve the highest scores. The only exceptions are in C1(Aircraft carrier) and C9(Non-ship), where IELT achieves the highest score in C1, while CLIP obtains the highest score in C9. For the confused categories C3, C5, and C6, the comparative methods have notably low reliability scores, with C5 even reaching negative values and C3 and C6 not exceeding 32\%. In contrast, IFShip(jud) maintains reliability scores above 60\% and IFShip(all) achieves scores no lower than 46\% in these three categories.

In \textcolor{blue}{Table \ref{tab:new}}, all methods show low reliability scores on the FGSCR42 dataset, with the OS score of all comparative methods being negative, and IFShip achieving only 34.05\%. However, IFShip maintains stable performance on C3, C5, and C6, avoiding negative scores seen in other methods. Notably, IFShip achieves 54.55\% on C6 while the best-performing method, VAN, achieves -39.23\%. Overall, IFShip demonstrates superior generalization compared to the comparative methods.

\textcolor{blue}{Table \ref{tab:class1}-\ref{tab:new}} show the highest accuracy and reliability of our method. This results from two reasons: 

Firstly, our IFShip trained with CoT instructions learns task-specific reasoning logic for FGSC. The multi-turn dialogue within CoT enables IFShip to progressively extract discriminative features from ship images under varying imaging conditions. This allows our IFShip to adaptively capture key features of different types of ships, leading to strong classification performance across categories. For example, in the results presented in \textcolor{blue}{Table \ref{tab:class1}} and \textcolor{blue}{\ref{tab:rel1}}, IFShip achieves high accuracy and reliability scores on the often-confused categories C3, C5, and C6. In contrast, the comparative methods, lacking such reasoning logic, suffer from severe confusion among these categories, resulting in low accuracy and reliability scores.

Secondly, IFShip benefits from domain knowledge in textual form during training. This enhancement enables IFShip to semantically align discriminative visual features from ship images with textual knowledge, thereby inferring fine-grained categories based on visible features. This differs from end-to-end methods, which directly classify ship categories based solely on visual features. Although CLIP also utilizes brief textual labels during training and classification phase, the support provided by these textual labels is evidently limited when compared to the rich domain knowledge in the TITANIC-FGS dataset. As a result, IFShip exhibits the best generalization. For instance, as shown in \textcolor{blue}{Table \ref{tab:new}}, despite being trained on drone-view images from the TITANIC-FGS dataset, IFShip maintains strong classification accuracy on satellite images in the FGSCR42 dataset, with superior reliability scores across most categories. In contrast, the comparative methods, relying solely on visual features, make severe classification errors, with some categories achieving 0\% accuracy.

We also observed that although CLIP follows the VLM architecture, it exhibits inferior performance compared to IFShip and even some end-to-end methods. We identified the following two main reasons: Firstly, the information learned by CLIP from label texts is insufficient. In FGSC task, focusing on specific component features of ship targets is crucial, which is a consensus within the FGSC domain and the key idea behind the TITANIC-FGS dataset. However, CLIP only has access to textual labels during training phase, which reflect global features of the ship target, rather than more detailed, component-specific information. As a result, CLIP primarily focuses on global features, limiting its classification performance. This leads to CLIP outperforming some end-to-end classification models (such as DCL and TASN) that are less effective at learning local features. However, it still lags behind IFShip and Acc-ViT, which excel in capturing and learning local features. Secondly, the dual-tower architecture of CLIP limits the interaction between ship images and textual labels to the embedding space, lacking finer-grained information exchange. While this may not be significant in general classification tasks, it is critical in FGSC task, where the visual features of different ships can be very similar, making the differences in similarity calculations minimal. In such cases, it is essential to jointly align image and text features for FGSC task. This is evident in the results on the C9 (non-ship) samples, where the IFShip model achieves 100\% accuracy in determining the presence of a ship target based on task-specific information, while CLIP only achieves 72.01\%.

Furthermore, we find that the reliability score designed in this paper provides a distinct perspective for evaluating model performance, different from accuracy. From the results in \textcolor{blue}{Table \ref{tab:class1}} and \textcolor{blue}{\ref{tab:rel1}}, we find that IELT does not achieve the highest classification accuracy in C1 and C9, but its reliability scores outperform all other methods. The reason is that IELT is confident in correct classifications, while errors stem from ambiguous, uncertain predictions. This results in high positive scores for correct samples and low penalty scores for misclassified ones. The phenomenon is also observed in the experimental results of CLIP in C9. Similarly, while IFShip(all) has lower accuracy than IFShip(jud), their reliability scores are similar. Both methods score all test images, but IFShip(jud) assigns a score of 0 to the 398 unclassified images, while IFShip(all) classifies and scores these samples normally. Since the ratio of incorrect to correct classifications for IFShip(all) on this subset is close to 1:1, the OS scores of IFShip(all) and IFShip(jud) are very similar. This phenomenon underscores the necessity of our proposed reliability score.

\subsection{Results of Interpretability for the FGSC Task}
\label{subsec4-4}
We evaluate the interpretability of advanced open-source VLMs and our IFShip model through two sub-tasks: ship image caption and ship image VQA. The overall accuracy (OA) of each VLM on the two tasks are shown in \textcolor{blue}{Table \ref{tab:conv1}}.

\begin{table}[h]
  \centering
  \huge
  \caption{Comparative results of different VLMs on image caption task and image VQA task.}
  \renewcommand{\arraystretch}{0.9} 
  \resizebox{0.8\columnwidth}{!}{
    \begin{tabular}{c|cc|cc}
    \hline
    \multirow{2}{*}{Model} & \multicolumn{2}{c|}{\makecell[c]{Fine-grained Ship Image Caption}} & \multicolumn{2}{c}{\makecell[c]{Fine-grained Ship Image VQA}} \\ \cline{2-5}
                                & GPT-3.5 & GPT-4.0 & GPT-3.5 & GPT-4.0 \\ \hline
    LLaVA\cite{cite18}  & 30.98   & 28.64   & 20.98   & 24.44   \\
    MoE-LLaVA(Phi-2)\cite{cite25}  & 39.80   & 20.65   & 21.32   & 29.89    \\
    MoE-LLaVA(Qwen)\cite{cite25}   & 49.05   & 17.49   & 22.34   & 19.35     \\
    MiniGPT-4(Llama2)\cite{cite24} & 52.95   & 27.03   & 10.40   & 19.20     \\
    MiniGPT-4(Vicuna0)\cite{cite24} & 58.69  & 26.16   & 11.46   & 24.86      \\
    Bunny\cite{cite26}          & 70.73   & 32.44   & 23.77   & 12.91   \\
    InstructBLIP\cite{cite27}      & 57.77   & 23.72   & 22.54   & 28.61   \\
    GeoChat\cite{cite69}      & 48.22   & 19.97   & 10.80   & 15.85   \\ \hline
    IFShip(Ours)            & \textbf{92.01}   & \textbf{76.91}    & \textbf{32.42}   & \textbf{40.12}   \\ \hline
    \end{tabular}}
    \label{tab:conv1}
\end{table}

\begin{figure*}[ht]
  \centering
   \includegraphics[width=0.96\linewidth]{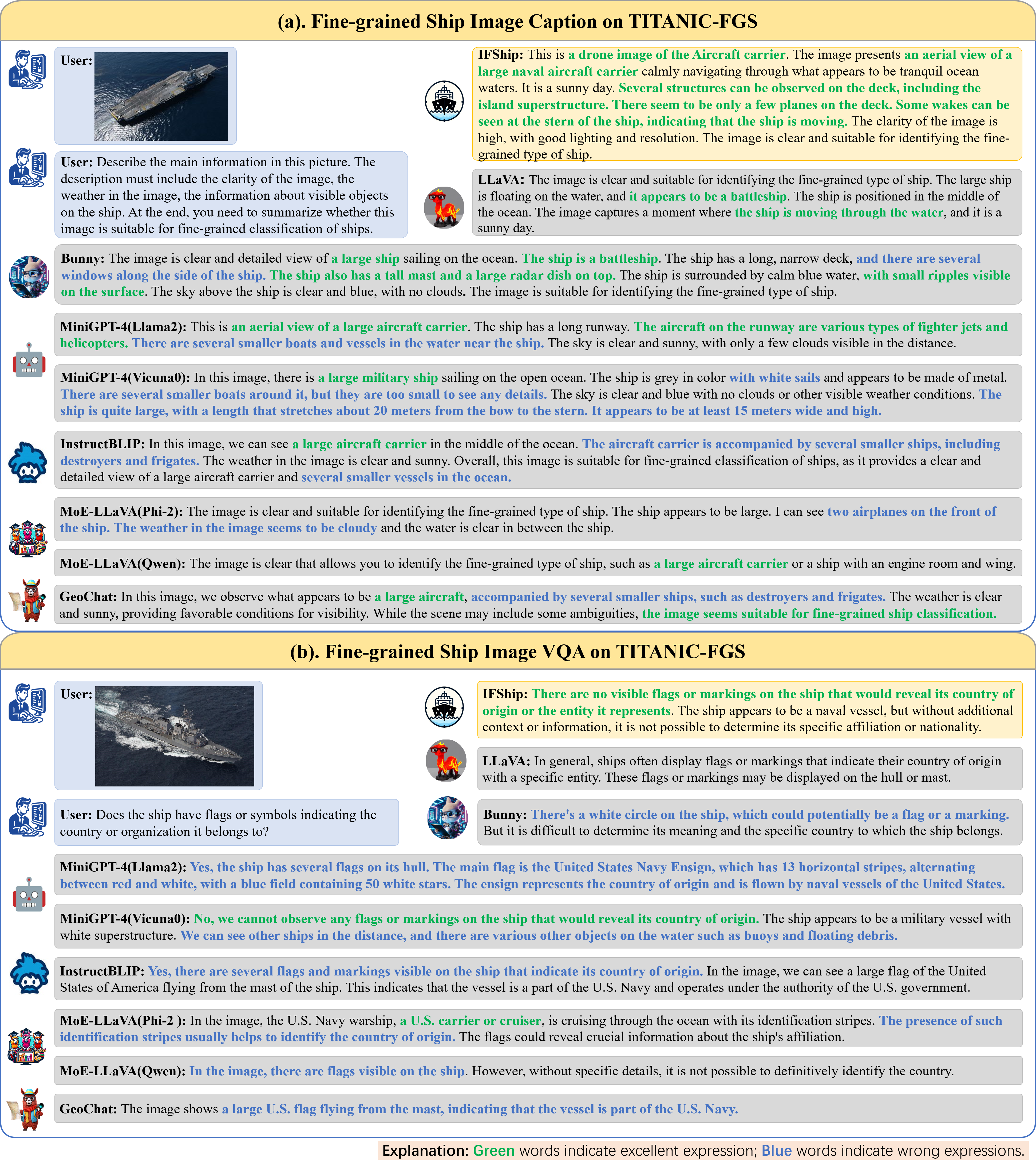}
   \caption{(a) Qualitative comparison results of different VLMs on the Ship Image Caption task. (b) Qualitative comparison results of different VLMs on the Ship Image VQA task.}
   \label{fig:conversation example}
\end{figure*}

\textbf{\textit{Fine-grained Ship Image Caption:}} The ship image caption task is crucial for evaluating the model's ability to summarize key information in ship images, because the FGSC task primarily relies on this information. For this task, the model generates descriptions that meet the requirements based on input ship images and textual task instructions. From the results in \textcolor{blue}{Table \ref{tab:conv1}}, IFShip achieves outstanding performance, with scores of 92.01\% and 76.91\% under the evaluation of the expert models GPT-3.5 and GPT-4.0, respectively. Compared to the second-best VLM, Bunny, our IFShip achieves a lead of 21.28\% and 44.47\%. Even compared to GeoChat, which was also pre-trained on remote sensing images, IFShip demonstrates an improvement of 43.79\% and 56.94\%. \textcolor{blue}{Fig. \ref{fig:conversation example} (a)} displays some qualitative results of the ship image caption task comparison. These results confirm that IFShip shows strong ship information comprehension capabilities after domain knowledge-enhanced instruction tuning. It can effectively explain the reasons for determining the corresponding ship categories and provide accurate evidence to assist users in assessing the reliability of the results. In contrast, MoE-LLaVA and GeoChat generate descriptions with significant content gaps, missing most required information. The LLaVA model offers relatively richer content but lacks clarity in its expression. While MiniGPT-4 and InstructBLIP produce detailed content, they are prone to errors and irrelevant conclusions (e.g., misidentifying waves as small boats), leading to poor performance. Bunny, the second-best model, excels in content richness and accuracy but fails to express detailed ship target information, resulting in lower performance than IFShip.

\textbf{\textit{Fine-grained Ship Image VQA:}} The ship image VQA task requires the VLM to extract relevant information and organize it into answers based on specific questions. According to the results in \textcolor{blue}{Table \ref{tab:conv1}}, IFShip still performs the best among all VLMs with accuracy of 32.42\% and 40.12\%, leading the second-best model by 8.65\% and 10.23\%, respectively. Combining with the examples shown in \textcolor{blue}{Fig.\ref{fig:conversation example} (b)}, we can easily observe that other VLMs, such as InstructBLIP and GeoChat, exhibit significant hallucinations when answering questions, even when no flags are present in the image. MiniGPT-4(Llama2) even incorrectly describes specific details, such as the number of stars on a flag. Although Bunny does not suffer from hallucinations, it tends to focus on irrelevant details, indicating biases in ship information perception. In contrast, after FGSC domain knowledge-enhanced instruction tuning, IFShip demonstrates a more precise and accurate understanding of the concept, effectively avoiding such issues.

From the results in \textcolor{blue}{Table \ref{tab:conv1}} and \textcolor{blue}{Fig.\ref{fig:conversation example}}, we make the following three observations:

Firstly, through CoT instruction tuning, our IFShip provides more comprehensive descriptions of ship images and more accurate ship category predictions. As demonstrated in the ship image caption task, IFShip generates descriptions that meet the instruction requirements. In contrast, other VLMs produce incomplete or erroneous descriptions due to insufficient understanding of concepts such as ship components or image clarity, often resulting in logical inconsistencies and lower accuracy.

Secondly, the incorporation of domain knowledge enhances IFShip's understanding of ship components, leading to accurate responses without hallucinations. This is particularly evident in \textcolor{blue}{Fig.\ref{fig:conversation example} (b)}, where many VLMs produce hallucinations in the absence of flags in the image. These hallucinations stem from a lack of understanding of ship flag targets. In terms of explanation reliability, the contrast between IFShip and LLaVA is striking. LLaVA offers general suggestions, such as the possibility of a flag indicating the ship's country or organization, without directly analyzing the input image. This vague response likely results from its limited domain knowledge. In contrast, IFShip delivers specific and precise descriptions, directly aligned with the image content and the explanation provided is also highly reasonable.

Finally, model performance on a specific task is not solely determined by parameter size; the role of fine-tuning is critical. Although our IFShip model is smaller in terms of model size and training data compared to some public VLMs, it significantly outperforms these methods in ship Image Caption and Image VQA tasks, thanks to the carefully designed domain knowledge-enhanced chain-of-thought instructions. In contrast, the largest model MiniGPT-4 (Vicuna0) performs moderately in the Image Caption task and near the bottom in the Image VQA task. The 3B-parameter Bunny model excels in the Image Caption task and performs well in the Image VQA task. Similarly, smaller MoE-LLaVA models outperform some 7B-parameter models across several metrics. This indicates that while the VLMs we selected are not excessively large, the key to performance lies in training methods and data sources, rather than simply increasing parameter size.

\subsection{Ablation Studies}
\label{subsec4-5}
In this section, we explore the impact of domain knowledge-enhanced instructions on model performance improvement. We refer to the IFShip model trained with domain knowledge-enhanced CoT instructions as {IFShip\textsubscript{CoT}}. Furthermore, we refer to the IFShip model trained with the instructions based on the template-filling method from \textcolor{blue}{Fig. \ref{fig:method compared} (a)} as {IFShip\textsubscript{NoCoT}}. The two models are trained on the same image data, but with different instruction data. We will discuss the effectiveness of domain knowledge-enhanced instruction construction method by comparing the performance of the two models. The experimental results are shown in \textcolor{blue}{Table \ref{tab:class2}-\ref{tab:conv2}}. For the FGSC task, we use the same experimental setup as in Section 4.3, comparing both models' performance under the (jud) and (all) settings. However, for the FGSCR42 dataset, both models classify all samples, so no distinction between the two settings is made.

\begin{table*}[ht]
  \centering
  \huge
  \caption{Accuracy comparative results of {IFShip\textsubscript{CoT}} and {IFShip\textsubscript{NoCoT}} on TITANIC-FGS dataset.}
  \renewcommand{\arraystretch}{1} 
  \resizebox{1\columnwidth}{!}{
    \begin{tabular}{l|*{17}{c}|c}
    \hline
    \multirow{2}{*}{\textbf{Method}} & \multicolumn{17}{c|}{\textbf{AR}} & \multirow{2}{*}{\textbf{OA}} \\ \cline{2-18}
                         & \textbf{C1} & \textbf{C2} & \textbf{C3} & \textbf{C4} & \textbf{C5} & \textbf{C6} & \textbf{C7} & \textbf{C8} & \textbf{C9} & \textbf{C10} & \textbf{C11} & \textbf{C12} & \textbf{C13} & \textbf{C14} & \textbf{C15} & \textbf{C16} & \textbf{C17} \\ \hline
    {IFShip\textsubscript{NoCoT}(jud)}                & 82.35 & 87.01 & 50.89 & 45.56 & 67.50 & 59.05 & 60.25 & 82.86 & 82.57 & 80.31 & 58.75 & 61.07 & 74.70 & 78.45 & 61.21 & 80.00 & 81.01 & 68.75               \\
    {IFShip\textsubscript{CoT}(jud)}               & \textbf{97.25} &  \textbf{95.19} &  \textbf{89.13} & \textbf{86.21} & \textbf{91.20} & \textbf{90.10}  & \textbf{100.00}   & \textbf{89.69} & \textbf{100.00}   & \textbf{100.00}   & \textbf{95.45} & \textbf{88.54} & \textbf{97.50} & \textbf{100.00}   & \textbf{100.00} & \textbf{83.64} & \textbf{98.04} & \textbf{94.02}               \\ \hline
    IFShip\textsubscript{NoCoT}(all) & \textbf{81.55} & \textbf{87.01} & 50.44 & 45.56 & 64.80 & 59.05 & 60.25 & \textbf{82.86} & 82.57 & 80.47 & 58.75 & 61.07 & 74.70 & 78.45 & 61.21 & \textbf{80.00} & 81.01 & 68.53               \\
    IFShip\textsubscript{CoT}(all) & 71.10 & 75.79 & \textbf{80.70} & \textbf{84.85} & \textbf{78.44} & \textbf{85.84} & \textbf{100.00} & 79.13 & \textbf{100.00} & \textbf{94.26} & \textbf{86.81} & \textbf{89.29} & \textbf{95.45} & \textbf{96.19} & \textbf{94.12} & 76.98 & \textbf{93.58} & \textbf{85.44}               \\ \hline
    \end{tabular}}
    \label{tab:class2}
\end{table*}

\begin{table*}[ht]
  \centering
  \huge
  \caption{Reliability comparative results of {IFShip\textsubscript{CoT}} and {IFShip\textsubscript{NoCoT}} on TITANIC-FGS dataset(${\gamma}$=0.6).}
  \renewcommand{\arraystretch}{1} 
  \resizebox{1\columnwidth}{!}{
    \begin{tabular}{l|*{17}{c}|c}
    \hline
    \multirow{2}{*}{\textbf{Method}} & \multicolumn{17}{c|}{\textbf{SR}} & \multirow{2}{*}{\textbf{OS}} \\ \cline{2-18}
                         & \textbf{C1} & \textbf{C2} & \textbf{C3} & \textbf{C4} & \textbf{C5} & \textbf{C6} & \textbf{C7} & \textbf{C8} & \textbf{C9} & \textbf{C10} & \textbf{C11} & \textbf{C12} & \textbf{C13} & \textbf{C14} & \textbf{C15} & \textbf{C16} & \textbf{C17} \\ \hline
    {IFShip\textsubscript{NoCoT}(jud)}                & 21.74 & 60.48 & 0.88  & 67.35 & -3.55 & -6.06 & 60.66 & 68.93 & 62.16 & 68.03 & 49.21 & 46.79 & 21.36 & 51.67 & 43.00 & 21.05 & 20.75 & 37.36               \\
    {IFShip\textsubscript{CoT}(jud)}               & \textbf{75.36} & \textbf{57.49} & \textbf{69.03} & \textbf{74.49} & \textbf{60.36} & \textbf{61.36} & \textbf{46.72} & \textbf{83.50}  & \textbf{85.59} & \textbf{79.51} & \textbf{83.33} & \textbf{75.23} & \textbf{66.99} & \textbf{71.67} & \textbf{60.00} & \textbf{78.07} & \textbf{91.51} & \textbf{70.97}               \\ \hline
    IFShip\textsubscript{NoCoT}(all) & 21.74 & 60.48 & 0.88 & 67.35 & -4.14 & -6.06 & \textbf{59.02} & 68.93 & 62.16 & 68.85 & 49.21 & 46.79 & 20.39 & 51.67 & 42.00 & 19.30 & 20.75 & 37.07               \\
    IFShip\textsubscript{CoT}(all) & \textbf{78.26} & \textbf{72.46} & \textbf{62.83} & \textbf{71.43} & \textbf{55.03} & \textbf{46.97} & 22.95 & \textbf{76.70} & \textbf{85.59} & \textbf{88.52} & \textbf{98.41} & \textbf{83.49} & \textbf{63.11} & \textbf{68.33} & \textbf{60.00} & \textbf{87.72} & \textbf{92.45} & \textbf{70.87}              \\ \hline
    \end{tabular}}
    \label{tab:rel2}
\end{table*}

\begin{table*}[htbp]
  \centering
  \huge
  \caption{Accuracy and Reliability comparative results of IFShip\textsubscript{CoT} and IFShip\textsubscript{NoCoT} on FGSCR42 dataset(${\gamma}$=0.6).}
  \renewcommand{\arraystretch}{1} 
  \resizebox{1\columnwidth}{!}{
    \begin{tabular}{l|*{7}{c}|*{7}{c}|c|c}
    \hline
    \multirow{2}{*}{\textbf{Method}} & \multicolumn{7}{c|}{\textbf{AR}} & \multicolumn{7}{c|}{\textbf{SR}} & \multirow{2}{*}{\textbf{OA}} & \multirow{2}{*}{\textbf{OS}} \\ \cline{2-15}
                         & \textbf{C1} & \textbf{C3} & \textbf{C5} & \textbf{C6} & \textbf{C7} & \textbf{C10} & \textbf{C17} & \textbf{C1} & \textbf{C3} & \textbf{C5} & \textbf{C6} & \textbf{C7} & \textbf{C10} & \textbf{C17} & \\ \hline
    IFShip\textsubscript{NoCoT}         & 38.75 & 3.69 & 30.18 & 1.72 & 9.80 & 78.82 & 81.41 & \textbf{98.29} & -96.05 & -44.61 & -97.73 & -97.46 & -70.55 & -73.62 & 31.76 & -36.48             \\ \hline
    IFShip\textsubscript{CoT}               & \textbf{80.67} & \textbf{93.26} & \textbf{91.99} & \textbf{100.00} & \textbf{100.00} & \textbf{98.48} & \textbf{99.84} & 95.93 & \textbf{9.39} & \textbf{4.94} & \textbf{54.55} & \textbf{-69.54} & \textbf{99.12} & \textbf{32.50} & \textbf{67.03} & \textbf{34.05}             \\ \hline
    \end{tabular}}
    \label{tab:new1}
\end{table*}

From the experimental results in \textcolor{blue}{Table \ref{tab:class2}} and \textcolor{blue}{Table \ref{tab:rel2}}, IFShip\textsubscript{NoCoT} performs significantly lower in the FGSC task compared to {IFShip\textsubscript{CoT}}. In the image quality discrimination module, {IFShip\textsubscript{NoCoT}} excludes only 8 out of 2053 images in the testing dataset as samples are not suitable for the FGSC task. Therefore, the experimental results under {IFShip\textsubscript{NoCoT}(jud)} and {IFShip\textsubscript{NoCoT}(all)} settings are similar. But this phenomenon is unreasonable, as IFShip\textsubscript{NoCoT} incorrectly classifies many unsuitable images as suitable, leading to poor task performance. In terms of experimental results, the {IFShip\textsubscript{NoCoT}(jud)} falls behind {IFShip\textsubscript{CoT}(jud)} by 25.27\% in OA and 33.61\% in OS; the {IFShip\textsubscript{NoCoT}(all)} also falls behind {IFShip\textsubscript{CoT}(all)} by 16.91\% in OA and 33.80\% in OS. For AR and SR, {IFShip\textsubscript{NoCoT}(jud)} performs worse than {IFShip\textsubscript{CoT}(jud)} across all categories, while {IFShip\textsubscript{NoCoT}(all)} slightly outperforms {IFShip\textsubscript{CoT}(all)} in a few categories. This phenomenon is also evident in \textcolor{blue}{Table \ref{tab:new1}}, where {IFShip\textsubscript{NoCoT}} underperforming in all aspects compared to {IFShip\textsubscript{CoT}}, except for the SR in C1(Aircraft carrier). This is due to {IFShip\textsubscript{NoCoT}} classifying many samples as C1, which boosts the SR for C1, while scores for other categories approach -1. This further highlights that {IFShip\textsubscript{NoCoT}} has inferior task transfer ability compared to {IFShip\textsubscript{CoT}}.

\begin{table}[h]
  \centering
  \huge
  \caption{Comparative results of IFShip\textsubscript{CoT} and IFShip\textsubscript{NoCoT} on image caption task and image VQA task.}
  \renewcommand{\arraystretch}{1} 
  \resizebox{0.8\columnwidth}{!}{
    \begin{tabular}{c|cc|cc}
    \hline
    \multirow{2}{*}{Model} & \multicolumn{2}{c|}{\makecell[c]{Fine-grained Ship Image Caption}} & \multicolumn{2}{c}{\makecell[c]{Fine-grained Ship Image VQA}} \\ \cline{2-5}
                                & GPT-3.5 & GPT-4.0 & GPT-3.5 & GPT-4.0 \\ \hline
    {IFShip\textsubscript{NoCoT}}  & 32.00   & 24.84   & 16.32   & 21.10   \\
    {IFShip\textsubscript{CoT}} & \textbf{92.01}   & \textbf{76.91}   & \textbf{32.42}   & \textbf{40.12}   \\ \hline
    \end{tabular}}
    \label{tab:conv2}
\end{table}

From the experimental results in \textcolor{blue}{Table \ref{tab:conv2}}, we observe that compared to {IFShip\textsubscript{CoT}}, {IFShip\textsubscript{NoCoT}} exhibits accuracy declines of 60.01\% and 52.07\% in the ship image caption task. For ship image VQA task, {IFShip\textsubscript{NoCoT}} exhibits accuracy declines of 16.10\% and 19.02\%. These results are quite surprising, as {IFShip\textsubscript{NoCoT}}'s performance on the two tasks is worse than some general-domain VLMs. 

The above experimental results validate the approach proposed in this paper: incorporating domain knowledge to construct CoT instructions for fine-tuning VLMs significantly enhances task performance. Although the training data is the same, the differences in instruction formats lead to a substantial performance gap between {IFShip\textsubscript{CoT}} and {IFShip\textsubscript{NoCoT}}. The descriptions related to domain knowledge can offer additional conceptual details, and the principles of CoT instruction generation can provide complete logical chains for instructions. However, the template-filling method lacks these two key elements, results in instructions that fail to help {IFShip\textsubscript{NoCoT}} capture the critical features of ship targets and prevents the model from learning task-specific logic. As a result, {IFShip\textsubscript{NoCoT}} performs even worse than end-to-end classification methods on the FGSC task, and its performance in the Image Caption and Image VQA tasks is even lower than that of untrained VLMs.

\section{Conclusion}
\label{subsec5}
This paper introduces a novel domain knowledge-enhanced CoT prompt generation mechanism, which integrates task-specific domain knowledge and execution logic into multi-round dialogue instructions. Based on this mechanism, we construct TITANIC-FGS, the first instruction-following dataset specifically designed for the FGSC task. Through fine-tuning on this dataset, we have developed IFShip, a VLM specialized for the FGSC task based on the LLaVA framework. Compared to general VLMs such as LLaVA and MiniGPT-4, IFShip shows improved performance by avoiding incorrect answers and hallucinations. Based on IFShip, we develop an FGSC visual chatbot that overcomes the interpretability limitations of traditional end-to-end classification methods. Extensive experiments demonstrate IFShip’s superior performance, surpassing VLMs like LLaVA and GeoChat by avoiding hallucinations and incorrect answers, underscoring the effectiveness and potential of domain knowledge-enhanced instruction tuning for the FGSC task. However, we acknowledge the limitations of our current work and identify several promising directions for future research. Firstly, the deeper exploration of the interpretability is crucial to further enhancing model transparency and understanding. Secondly, we plan to investigate model architectures that are more tailored to fine-grained classification tasks, with a focus on optimizing VLMs for such specialized applications. Finally, we aim to extend our approach to other fine-grained classification tasks, examining its adaptability and performance across different domains. These directions will guide our future efforts in advancing the field.

 \bibliographystyle{IEEEtran}
 
\end{document}